\documentclass[11pt]{article}

\usepackage[preprint]{acl}

\usepackage{times}
\usepackage{latexsym}
\usepackage[T1]{fontenc}
\usepackage[utf8]{inputenc}
\usepackage{microtype}
\usepackage{inconsolata}
\usepackage{graphicx}
\usepackage{amsmath}
\usepackage{amssymb}
\usepackage{booktabs}
\usepackage{multirow}
\usepackage{xcolor}
\usepackage{url}
\usepackage{algorithm}
\usepackage{algpseudocode}
\usepackage{hyperref}
\usepackage{dsfont}

\newcommand{\Ncyc}{N_{\mathrm{cyc}}}
\newcommand{\jsd}{\mathrm{JSD}}
\newcommand{\kl}{\mathrm{KL}}

\newcommand{\HH}{\mathcal{H}}

\title{Uncertainty Decomposition via Cyclical SG-MCMC and Soft-label Learning for Subjective NLP}

\author{Keito Inoshita\textsuperscript{1,2,*} \quad Takato Ueno\textsuperscript{3} \\
  \textsuperscript{1}Faculty of Business and Commerce, Kansai University, Suita, Japan \\
  \textsuperscript{2}Data Science and AI Innovation Research Promotion Center, Shiga University, Hikone, Japan \\
  \textsuperscript{3}Graduate School of Data Science, Shiga University, Hikone, Japan \\
  \texttt{inosita.2865@gmail.com} \quad \texttt{s7025101@st.shiga-u.ac.jp}}

\begin{document}
\maketitle

\begin{abstract}
Annotator disagreement in emotion classification reflects ambiguity intrinsic to emotion concepts and is essential for predictor-quality assessment in subjective NLP. Yet no prior work integrates soft-label learning with Bayesian deep learning to evaluate uncertainty along axes including annotator-distribution fidelity. We train a linear head on a frozen RoBERTa via cyclical stochastic gradient Markov chain Monte Carlo (cSG-MCMC), targeting the empirical annotator distribution with a soft-label objective under a five-axis evaluation. On the $28$-emotion GoEmotions benchmark, the proposed method outperforms Monte Carlo Dropout and Deep Ensemble simultaneously on three axes---Jensen-Shannon divergence (JSD) to the annotator distribution, Spearman correlation between per-emotion aleatoric uncertainty and disagreement, and selective-prediction Area Under the Risk-Coverage Curve (AURC) and Area Under the ROC Curve (AUROC)---showing independent axes are jointly attainable from one posterior. Post-hoc temperature scaling exhibits a bidirectional effect, establishing hard-label calibration and annotator-JSD as independent dimensions and motivating joint reporting as an honest protocol.
\end{abstract}

\section{Introduction}\label{sec:intro}

Annotator disagreement in emotion classification is not noise but a structure that reflects an ambiguity intrinsic to emotion concepts themselves. \citet{plank2022} and \citet{frenda2024perspectivist} consolidate that annotator disagreement appears systematically as an inherent structure across many natural language processing (NLP) tasks, and argue for preserving perspectives across data, model, and evaluation. Mainstream NLP evaluation, however, relies on majority-vote hard labels and discards this structure. For subjective NLP tasks, we therefore need to revisit, at a methodological level, how faithfully a predictor can represent the annotator distribution \citep{keito2026bridge}.

Two research threads address this in parallel. The first is soft-label learning (\citealp{peterson2019humanuncertainty}; \citealp{uma2021}), which formulates objectives pushing the predictive distribution towards the empirical annotator distribution. The second is Markov chain Monte Carlo (MCMC) heads in Bayesian deep learning, cyclical stochastic gradient MCMC (cSG-MCMC, \citealp{zhang2020csgmcmc}); Monte Carlo Dropout (MC Dropout, \citealp{gal2016dropout}); Deep Ensemble \citep{lakshminarayanan2017}---decomposing posterior-derived uncertainty into aleatoric (data-intrinsic ambiguity) and epistemic (lack of model knowledge) components \citep{kendall2017}. No prior work combines the two threads to evaluate uncertainty quality and annotator-distribution alignment jointly in subjective NLP. We fill that gap and report what emerges empirically from this combined design.

A central difficulty in uncertainty evaluation for subjective NLP is that there is no settled definition of what makes uncertainty ``high quality''. Hard-label calibration metrics such as Brier score \citep{brier1950verification} and Expected Calibration Error (ECE; \citealp{guo2017oncalibration}) penalise mass spread away from the argmax and are structurally unfavourable to predictors faithful to the annotator distribution \citep{baan2024interpreting}. Beyond these, selective prediction \citep{geifman2017selective}, misclassification detection \citep{hendrycks2017baseline}, and per-category correlation provide complementary axes. We present a five-axis framework that jointly captures these dimensions: (C1) hard-label argmax calibration; (C2) Jensen-Shannon divergence (JSD) against the annotator distribution; (C3) the Spearman correlation between per-emotion aleatoric uncertainty and the per-emotion annotator disagreement rate; (C4) acquisition quality in active learning---a natural context for Bayesian Active Learning by Disagreement (BALD; \citealp{houlsby2011bald}); and (C5) selective prediction via the Area Under the Risk-Coverage Curve (AURC) and misclassification detection via the Area Under the ROC Curve (AUROC). The central question becomes whether uncertainty captures task-intrinsic ambiguity, whether it localises prediction errors, and whether accuracy-centric and distribution-centric calibration are independent dimensions.

We propose a predictor placing a cSG-MCMC head on a frozen RoBERTa backbone \citep{liu2019roberta} trained with a soft-label cross-entropy objective, and---under the five-axis framework of \S\ref{sec:method}---show what this design reveals on the $28$-emotion GoEmotions task \citep{demszky2020}. Concretely, the proposed method consistently improves over baselines in annotator-distribution JSD and selective-prediction AURC / AUROC; per-emotion aleatoric uncertainty correlates with annotator disagreement; and post-hoc temperature scaling \citep{guo2017oncalibration} moves accuracy-centric and distribution-centric calibration in opposite directions. We report hyperparameter dependence and position all main claims under a canonical setting.

\section{Related Work}\label{sec:related}

\subsection{Annotator disagreement and soft-label learning}\label{sec:rel_soft}

The treatment of annotator disagreement in NLP has shifted from ``aggregating it away as noise'' to ``preserving it as signal''. \citet{plank2022} and \citet{pavlick2019} redefine human label variation as a problem that spans data, models, and evaluation, and \citet{frenda2024perspectivist} methodologically establishes perspectivist NLP. On the learning side, building on the temperature-scaled soft-target distillation of \citet{hinton2015distilling}, \citet{peterson2019humanuncertainty} demonstrates robustness gains from learning the annotator distribution in image classification, and \citet{uma2021} organises the evaluation of soft labels. On the evaluation side, \citet{baan2022stop,baan2024interpreting} identify a structural problem in which conventional calibration conflates ``model confidence'' with ``human label variation''. On the dataset side, designs that preserve $3$--$5$ annotations per example, such as GoEmotions \citep{demszky2020}, have become more common, and the annotator-centric active learning of \citet{vandermeer2024annotator} likewise reflects a trend that handles the annotator distribution explicitly.

The main focus of these studies, however, is either on training with soft labels as the target or on evaluating soft labels, and the perspective of evaluating the predictor's uncertainty itself against the annotator distribution remains underdeveloped. We fill this gap by integrating soft-label learning with Bayesian uncertainty decomposition and by evaluating uncertainty along axes that include fidelity to the annotator distribution.

\subsection{Bayesian deep learning and MCMC heads}\label{sec:rel_bdl}

Uncertainty quantification in Bayesian deep learning has progressed from deterministic approximations (variational inference, the Laplace approximation) to MCMC-based approaches that directly handle the multi-modality of the posterior. Building on the stochastic gradient Langevin dynamics (SGLD) of \citet{welling2011sgld} and the SG-MCMC of \citet{chen2014sgmcmc}, \citet{zhang2020csgmcmc} proposes cSG-MCMC, which efficiently explores a multi-modal posterior via a cyclical step-size schedule and underlies our sampler. As lightweight approximations, MC Dropout \citep{gal2016dropout} and Deep Ensemble \citep{lakshminarayanan2017}---a canonical method for epistemic uncertainty---are widely adopted. The cold-posterior discussion is consolidated in \citet{wenzel2020,aitchison2021}, linear probing as a means of protecting pre-trained features in \citet{kumar2022finetuning}, and the formalism of aleatoric / epistemic decomposition is re-examined by \citet{wimmer2023quantifying,sale2024labelwise}.

These prior methods, however, are validated almost exclusively on image classification or hard-label evaluation, and we are not aware of any prior study that systematically asks how posterior-derived uncertainty corresponds to the shape of the annotator distribution in a subjective NLP task. We fill that gap by applying cSG-MCMC to the linear head of a frozen RoBERTa and evaluate aleatoric / epistemic uncertainty including alignment with the annotator distribution.

\subsection{Multi-axis evaluation of uncertainty}\label{sec:rel_uq}

Uncertainty evaluation in NLP has diversified in both methods and viewpoints. For aleatoric / epistemic decomposition, the mutual information formalism of \citet{kendall2017} is widely cited as a standard. On the calibration side, post-hoc temperature scaling \citep{guo2017oncalibration} continues to serve as a baseline. In the selective-prediction line, \citet{geifman2017selective} establishes the reject-and-defer formalism and AURC for deep networks, and for acquisition design in active learning, BALD \citep{houlsby2011bald} and BatchBALD \citep{kirsch2019batchbald} serve as standard methods.

While each of these metrics is mature in isolation, few studies integrate them to ask which metrics multi-facetedly capture uncertainty quality in subjective NLP. This paper presents an evaluation framework that applies, in parallel to a single predictor, the five axes of (C1) hard-label calibration, (C2) annotator JSD, (C3) the per-category correlation between aleatoric uncertainty and disagreement rate, (C4) active-learning acquisition, and (C5) selective-prediction AURC / AUROC, and within this framework, characterises the phenomenon that post-hoc temperature scaling moves accuracy-centric and distribution-centric calibration in opposite directions. We also report, as empirical evidence, the class-balance failure mode of BALD acquisition observed under the long-tail imbalance of GoEmotions.

\section{Method}\label{sec:method}

\subsection{Notation and dataset}\label{sec:notation}

Each example consists of an input text $x_i$ and labels $y_i^{(1:A_i)}$ over $C$ emotion categories provided by $A_i$ annotators, and we denote the full dataset as $\mathcal{D} = \{(x_i, y_i^{(1:A_i)})\}_{i=1}^{N}$. We construct two types of supervision: a majority-vote hard label $y_i^{\star} = \arg\max_c \sum_a \mathds{1}[y_i^{(a)}{=}c]$, and a soft label $q_i^{c} = \tfrac{1}{A_i} \sum_a \mathds{1}[y_i^{(a)}{=}c]$ that preserves the empirical annotator distribution. The specific dataset and splits are described in \S\ref{sec:setup}.

\subsection{Method definitions}\label{sec:methods-defs}

All four methods share a linear head on top of a frozen backbone. Freezing the backbone is a design choice that separates the comparison between pre-training quality and the uncertainty mechanism (Appendix~\ref{app:method_details}).

\paragraph{B0 (deterministic baseline).}
A single head with $M{=}1$ at inference. Epistemic uncertainty is structurally $0$, and B0 serves as an uncertainty-free reference point.

\paragraph{B1 (MC Dropout).}
The same configuration as B0 with dropout applied to the head input; multiple stochastic forward passes are taken at inference \citep{gal2016dropout}.

\paragraph{B2 (Deep Ensemble).}
Several independently initialised heads are trained with different data shuffles \citep{lakshminarayanan2017}.

\paragraph{Proposed method (cSG-MCMC head).}
We replace the head's optimiser with the cSG-MCMC sampler of \citet{zhang2020csgmcmc} and treat the resulting collection of weight samples as the posterior distribution. The step size follows a cosine schedule:
\begin{equation}
\alpha_t = \frac{\alpha_0}{2}
\left[ 1 + \cos\!\left(\frac{\pi \cdot \mathrm{mod}(t, K)}{K}\right) \right],
\label{eq:cosine}
\end{equation}
where $K$ is the cycle length. Each cycle is divided into two phases: the first $1{-}\xi$ portion is noise-free stochastic gradient descent (SGD; the exploration phase), and the final $\xi$ portion is SGLD with Gaussian noise (the sampling phase). Letting $L(\theta;\mathcal{B})$ denote the mini-batch mean cross-entropy and $p(\theta){=}\mathcal{N}(0,\sigma^2 I)$ the prior,
\begin{equation}
\begin{aligned}
\theta_{t+1}
={}& \theta_t - \alpha_t \bigl[ N \nabla_\theta L(\theta_t;\mathcal{B}_t)
       + \nabla_\theta(-\log p(\theta_t)) \bigr] \\
&{}+ \mathds{1}[\text{sampling}] \sqrt{2\alpha_t T}\,\epsilon_t,
\quad \epsilon_t \sim \mathcal{N}(0, I).
\end{aligned}
\label{eq:sgld}
\end{equation}
Here $T$ is the posterior temperature, $N{=}|\mathcal{D}|$ is the posterior scale, and the prior is implemented as L2 weight decay $\eta{=}1/\sigma^2$. We collect $S$ samples uniformly within the sampling phase of every cycle and, after $B$ cycles of burn-in, retain $M=(\Ncyc-B)\cdot S$ samples (Algorithm~\ref{alg:csgmcmc}). Here $\Ncyc$ denotes the number of cycles. Specific hyperparameter values ($\Ncyc$, $K$, $\xi$, $S$, $B$, $\alpha_0$, $T$, $\eta$) are given in \S\ref{sec:setup}.

\paragraph{Design rationale.}
The cosine schedule transitions smoothly within each cycle from large-step exploration to small-step sampling, allowing a single sampler to realise both inter-mode transitions of the multi-modal posterior and local sampling around each mode \citep{zhang2020csgmcmc}. The design of collecting samples only in the SGLD phase at the end of each cycle excludes the transient bias of exploration steps and secures the approximation accuracy of the posterior invariant distribution. Treating only the head in a Bayesian manner separates the comparison between pre-trained representations and the uncertainty mechanism, and is a deliberate design choice that maintains consistency with the frozen-backbone setting typical of practical NLP deployments.

\begin{algorithm}[t]
\small
\caption{cSG-MCMC head training}
\label{alg:csgmcmc}
\begin{algorithmic}[1]
\Require data $\mathcal{D}$, prior $p(\theta){=}\mathcal{N}(0,\sigma^2 I)$ (L2 weight decay $\eta{=}1/\sigma^2$), posterior temperature $T$, cycles $\Ncyc$, cycle length $K$, sampling-phase fraction $\xi$, samples per cycle $S$, burn-in $B$, posterior scale $N{=}|\mathcal{D}|$
\State $\theta_0 \sim p(\theta)$; \quad $\mathrm{Samples} \gets \emptyset$
\For{$t = 0, 1, \dots, \Ncyc K - 1$}
  \State $\alpha_t \gets$ Eq.~\eqref{eq:cosine}
  \State $\mathcal{B}_t \sim \mathcal{D}$
  \State $g_t \gets N \nabla_\theta L(\theta_t;\mathcal{B}_t)
                  + \nabla_\theta(-\log p(\theta_t))$
  \If{$(t \bmod K)/K < 1-\xi$}
    \State $\theta_{t+1} \gets \theta_t - \alpha_t g_t$ \Comment{exploration}
  \Else
    \State $\epsilon_t \sim \mathcal{N}(0, I)$
    \State $\theta_{t+1} \gets \theta_t - \alpha_t g_t
            + \sqrt{2\alpha_t T}\,\epsilon_t$ \Comment{SGLD sampling}
    \If{$\lfloor t/K \rfloor \ge B$ and $\mathrm{collect}(t)$}
      \State $\mathrm{Samples} \gets \mathrm{Samples} \cup \{\theta_t\}$
    \EndIf
  \EndIf
\EndFor
\State \Return $\mathrm{Samples}$ \Comment{$M = (\Ncyc - B)\cdot S$ samples}
\end{algorithmic}
\end{algorithm}

\subsection{Soft-label likelihood}\label{sec:soft-loss}

The soft-label loss is given by the Kullback-Leibler (KL) divergence $\kl(q_i \| p_\theta)$ from the annotator distribution $q_i$ to the model's predictive distribution $p_\theta(\cdot|x_i)$; up to the constant $-H(q_i)$ that does not depend on $\theta$, this is equivalent to maximising $\sum_c q_i^c \log p_\theta(c|x_i)$, which coincides with the soft-target mode of \texttt{F.cross\_entropy}. This is the same form of soft-target learning used by \citet{hinton2015distilling} and \citet{peterson2019humanuncertainty}. Under hard labels, $q_i$ degenerates to a one-hot vector. For fair comparison, the soft runs of B0 / B1 / B2 and the proposed method all share the same loss form. The significance of adopting the soft-label likelihood is that it preserves the shape of the empirical annotator distribution itself as the learning signal, securing a direct path from the annotator-level inherent disagreement contained in the vote distribution---which majority-vote hardening discards---to the model.

\subsection{Predictive distribution and uncertainty decomposition}\label{sec:uq}

From $M$ posterior samples $\{p_m(y|x)\}_{m=1}^{M}$, the predictive distribution is defined as the posterior mean $p_\theta(y|x) = \tfrac{1}{M} \sum_{m=1}^{M} p_m(y|x)$, and uncertainty is decomposed into total / aleatoric / epistemic components on this basis. Let $\mathcal{H}[p] = -\sum_y p(y) \log p(y)$ denote the Shannon entropy of a probability distribution $p$.
\begin{align}
\mathcal{H}_{\mathrm{tot}}(x) &= \mathcal{H}[p_\theta(\cdot|x)],
\label{eq:h_tot} \\
\mathcal{H}_{\mathrm{ale}}(x) &= \tfrac{1}{M} \sum_{m=1}^{M} \mathcal{H}[p_m(\cdot|x)],
\label{eq:h_ale} \\
\mathcal{H}_{\mathrm{epi}}(x) &= \mathcal{H}_{\mathrm{tot}}(x) - \mathcal{H}_{\mathrm{ale}}(x).
\label{eq:h_epi}
\end{align}
This decomposition follows the formalism of \citet{depeweg2018}: $\mathcal{H}_{\mathrm{tot}}(x)$ operationally defines total uncertainty as the entropy of the full predictive distribution, $\mathcal{H}_{\mathrm{ale}}(x)$ defines data-intrinsic ambiguity (aleatoric uncertainty) as the predictive entropy averaged over posterior samples, and $\mathcal{H}_{\mathrm{epi}}(x)$ defines the lack of model knowledge (epistemic uncertainty) as the disagreement among posterior samples (equivalent to the mutual information $I[y;\theta|x]$). For B0, the posterior samples collapse to a single point and $\mathcal{H}_{\mathrm{epi}}(x){=}0$ structurally. This decomposition serves as the direct basis for C3 and C5 in \S\ref{sec:c3_main} and \S\ref{sec:c5_main}.

\subsection{Evaluation metrics and statistical tests}\label{sec:metrics}

The main metrics are as follows: (C1) ECE / Brier score / negative log-likelihood (NLL) targeting the hard-label argmax; (C2) $\jsd(q\|p)$ against the annotator distribution (in bits, base $2$); (C3) the Spearman $\rho$ between the per-category mean aleatoric uncertainty $\bar{\HH}_{\mathrm{ale}}^{(c)}$ and the per-category annotator disagreement rate $\delta^{(c)}$; (C4) for the active-learning acquisition strategies $\in \{\text{BALD}, \text{entropy}, \text{random}\}$ \citep{houlsby2011bald}, we measure a learning curve of $20$ iterations $\times$ $500$ samples per iteration using the proposed method as the base predictor and assess acquisition quality with validation accuracy and macro-F1; (C5) selective-prediction AURC \citep{geifman2017selective} and misclassification-detection AUROC \citep{hendrycks2017baseline}. As sanity supplements, we also report accuracy and macro-F1. The disagreement rate $\delta^{(c)}$ is computed once from the original GoEmotions data and is identical across all runs (Appendix~\ref{app:method_details}).

For statistical testing, we report a two-way analysis of variance (ANOVA) with Type II sum of squares (SS), Holm--Bonferroni \citep{holm1979simple} corrected paired $t$-tests of the proposed method against each baseline, bootstrap $95\%$ confidence intervals (CIs; $1{,}000$ iterations, percentile), Cohen's $d$, and partial $\eta^2$. \emph{Strict dominance} is defined as the conjunction of (D1) paired differences signed in the direction of improvement across all seeds and (D2) Holm-corrected $p$-values for paired $t$-tests below $\alpha=0.05$. The full statistical protocol, backbone implementation, metric definitions, and details on strict dominance are collected in Appendix~\ref{app:method_details}.

\section{Experimental Setup}\label{sec:setup}

\subsection{Dataset and model}\label{sec:setup_data}

We use GoEmotions \citep{demszky2020}. This corpus takes Reddit comments as input, and each example has a vote vector over $C{=}28$ emotion categories. We adopt the canonical splits (train / validation / test $= 43{,}410 / 5{,}426 / 5{,}427$), and $A_i \in \{3, 4, 5\}$. Within the validation split, $1{,}544$ examples ($28.5\%$) form a high-disagreement subset to which a fifth annotator was added after the initial three disagreed; we use this subset for limit-case diagnostics of annotator-distribution fidelity (Appendix~\ref{app:5ann_subset}).

The model is a frozen RoBERTa-base \citep{liu2019roberta} (maximum length $128$, mean-pooled $[\mathrm{CLS}]$ representation fed into a linear head), shared across the four methods. The linear head maps $768 \to 28$ and contains $21{,}532$ trainable parameters. The number of members at inference is $M{=}1$ for B0, $M{=}20$ for B1 (dropout $p{=}0.1$), $M{=}5$ for B2, and $M{=}30$ for the proposed method.

\subsection{Training configuration and experimental protocol}\label{sec:setup_train}

For B0 / B1 / B2 we use AdamW (lr $10^{-3}$, weight decay $10^{-2}$, batch $32$, max $15$ epochs, early stopping with patience $3$ on validation NLL). The canonical cSG-MCMC setting for the proposed method is $\Ncyc{=}8$, $K{=}2{,}500$ (a total of $20{,}000$ steps), $\xi{=}0.25$, $S{=}5$, $B{=}2$ (hence $M{=}30$), $\alpha_0{=}10^{-4}$, $T{=}1.0$, and $\eta{=}10^{-4}$ (within $1.7\%$ of the B0 token budget of $20{,}355$). All main claims are established under this canonical setting.

All runs are executed on a single NVIDIA H100 NVL with \texttt{torch.use\_deterministic\_algorithms(True)} and bfloat16 mixed precision. Seeds are $\{42, 43, 44\}$, controlling PyTorch, NumPy, the random module, and the HuggingFace data loader in a unified manner.

Each method is run under both label modes (hard / soft), yielding a main grid of $4 \times 2 \times 3 = 24$ runs. In addition, we conduct $9$ active-learning runs, a robustness ablation, and post-hoc temperature scaling. Details are given in Appendix~\ref{app:setup_details}.

\section{Results}\label{sec:results}

\subsection{Hard-label calibration}\label{sec:results_c1}

On the test split we compare Brier multiclass and ECE targeting the hard-label argmax across all methods and both label modes (Table~\ref{tab:c1_brier_ece}). The proposed method shows larger values than baselines under both modes, attaining Brier $0.826$ / ECE $0.193$ under soft and Brier $0.847$ / ECE $0.248$ under hard; the best argmax calibration is Brier $0.772$ / ECE $0.061$ for B0 hard.

This underperformance is not a defect but a structural consequence: high-entropy predictions faithful to the annotator distribution are driven away from the argmax mass, and ECE / Brier penalise argmax concentration. The trade-off must be assessed on a dimension independent of the annotator JSD below.

\begin{table}[t]
\small
\centering
\caption{Brier and ECE with the hard-label argmax as target.}
\label{tab:c1_brier_ece}
\begin{tabular}{ll cc}
\toprule
Method & Label & Brier & ECE \\
\midrule
B0 & hard & $0.772 \pm 0.002$ & $0.061 \pm 0.011$ \\
B0 & soft & $0.788 \pm 0.002$ & $0.085 \pm 0.008$ \\
B1 & hard & $0.773 \pm 0.002$ & $0.077 \pm 0.015$ \\
B1 & soft & $0.790 \pm 0.002$ & $0.104 \pm 0.019$ \\
B2 & hard & $0.763 \pm 0.001$ & $0.080 \pm 0.024$ \\
B2 & soft & $0.779 \pm 0.001$ & $0.107 \pm 0.022$ \\
proposed & hard & $0.847 \pm 0.004$ & $0.248 \pm 0.005$ \\
proposed & soft & $0.826 \pm 0.007$ & $0.193 \pm 0.012$ \\
\bottomrule
\end{tabular}
\end{table}

\subsection{JSD against the annotator distribution}\label{sec:c2_main}
\label{sec:results_main}

\begin{figure}[t]
\centering
\includegraphics[width=0.95\columnwidth]{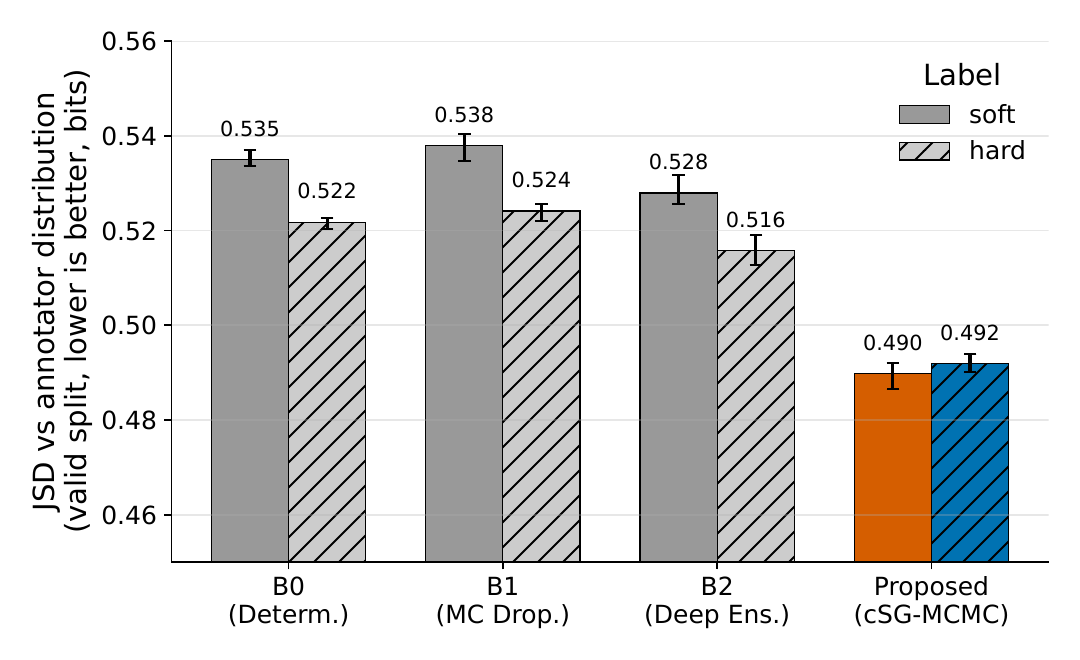}
\caption{JSD between the predictive distribution and the empirical annotator distribution.}
\label{fig:c2_jsd}
\end{figure}

Figure~\ref{fig:c2_jsd} shows the JSD (bits) between predictive and empirical annotator distributions on validation, across $4$ methods $\times$ $2$ labels. The proposed method attains JSD $0.490 \pm 0.003$ under soft, strictly dominating the best baseline B2 soft ($0.528 \pm 0.003$) across all $3$ seeds (Holm-corrected paired $t$, $p < 0.01$), as well as B0 soft ($0.535$) and B1 soft ($0.538$). Under hard, the proposed method ($0.492$) also consistently undercuts the baselines ($0.516$--$0.524$).

This strict dominance shows the posterior mean is geometrically closer to the empirical annotator distribution---a success in reproducing inter-annotator opinion structure that majority-vote hard evaluation cannot see, and evidence that distribution-centric evaluation is a significant independent verification axis for subjective NLP.

\subsection{Correlation between aleatoric uncertainty and annotator disagreement}\label{sec:c3_main}

Table~\ref{tab:c3_per_emotion} shows per-emotion Spearman $\rho$ between mean aleatoric uncertainty and annotator disagreement rate over $K = 23$ categories on validation. The proposed method attains $\rho = 0.489 \pm 0.030$ (soft) and $0.556 \pm 0.031$ (hard), exceeding the soft-baseline seed-mean of $\le 0.290$ (at B2) by more than $0.20$. The same ordering holds under hard, with bootstrap $95\%$ CIs of $[0.10, 0.78]$ (soft) and $[0.14, 0.84]$ (hard), both strictly above $0$. On test the values decay (soft $0.336$, hard $0.417$) but the qualitative ordering (proposed $>$ baselines) is preserved.

This correlation shows the proposed method's aleatoric uncertainty captures task-intrinsic ambiguity, functioning as a category-level interpretable signal specific to subjective NLP and distinguishable from epistemic uncertainty due to hallucination or out-of-distribution (OOD) inputs.

\begin{table}[t]
\small
\centering
\setlength{\tabcolsep}{4pt}
\caption{Per-emotion Spearman $\rho$ between aleatoric uncertainty and annotator disagreement rate.}
\label{tab:c3_per_emotion}
\begin{tabular}{ll cc}
\toprule
Method & Label & $\rho$ mean$\pm$std & boot. $95\%$ CI \\
\midrule
B0 & soft & $0.209 \pm 0.010$ & --- \\
B1 & soft & $0.250 \pm 0.018$ & --- \\
B2 & soft & $0.290 \pm 0.060$ & --- \\
proposed & soft & $0.489 \pm 0.030$ & $[0.10, 0.78]$ \\
\midrule
B0 & hard & $0.273 \pm 0.045$ & --- \\
B1 & hard & $0.308 \pm 0.023$ & --- \\
B2 & hard & $0.359 \pm 0.088$ & --- \\
proposed & hard & $0.556 \pm 0.031$ & $[0.14, 0.84]$ \\
\bottomrule
\end{tabular}
\end{table}

\subsection{Acquisition quality in active learning}\label{sec:results_c4}

\begin{figure*}[t]
\centering
\includegraphics[width=0.85\textwidth]{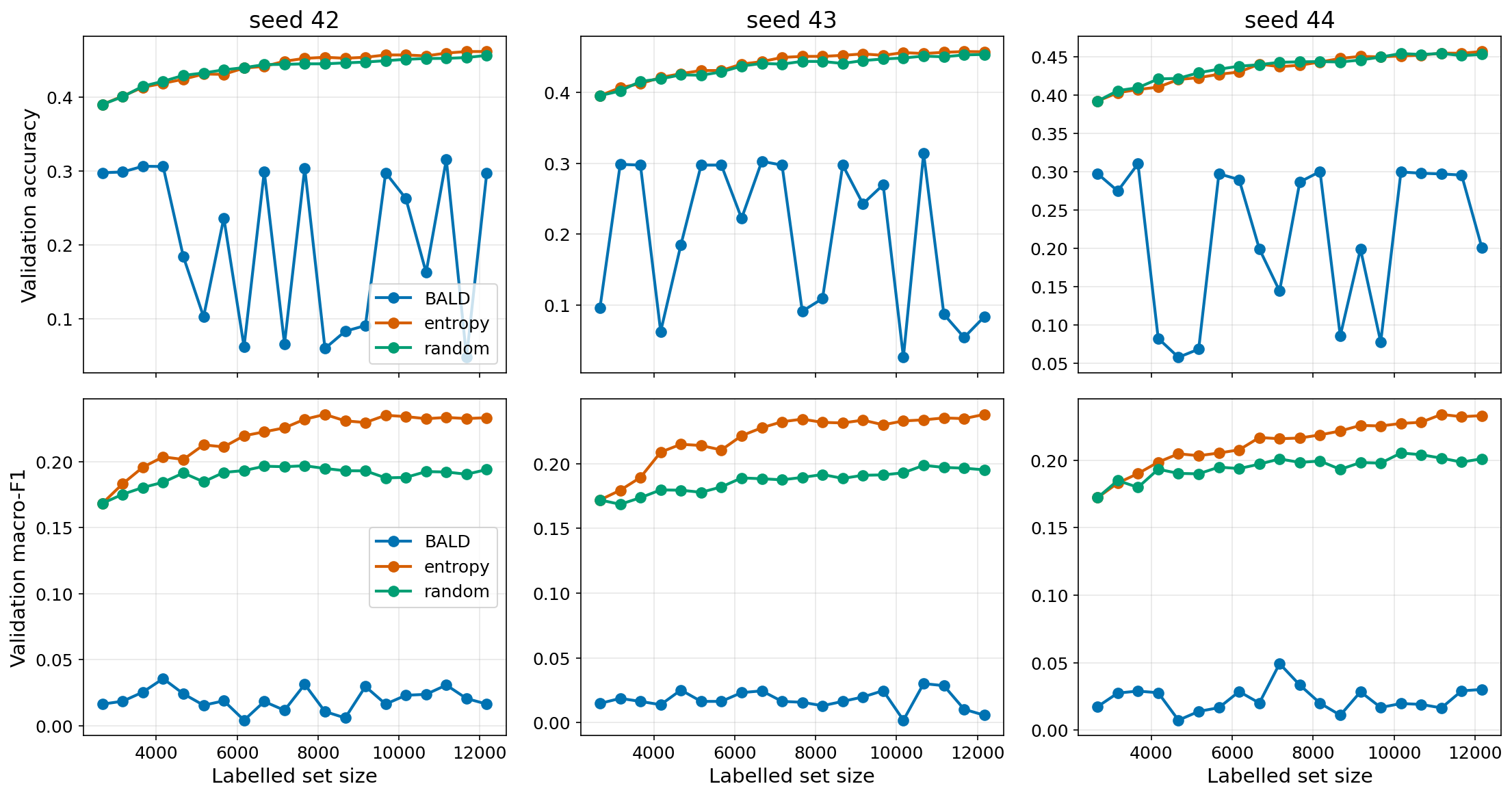}
\caption{Active-learning curves.}
\label{fig:al_curves}
\end{figure*}

Figure~\ref{fig:al_curves} shows active-learning curves using the proposed method's posterior-derived uncertainty as the base predictor. We ran $9$ total runs ($3$ acquisition strategies---BALD, predictive entropy, random---$\times$ $3$ seeds) and compared final-iteration validation accuracy. Entropy converges stably at $\{0.461, 0.457, 0.457\}$ and random at $\{0.456, 0.453, 0.454\}$, while BALD collapses across all $3$ seeds at $\{0.297, 0.084, 0.201\}$.

BALD's collapse versus the stability of entropy and random shows uncertainty reliability along the acquisition axis must be verified independently of the rejector axis (C5). Under GoEmotions' long-tail distribution, naive BALD without batch-level diversity exhibits structural over-concentration on rare classes, while entropy-based acquisition functions robustly despite its simplicity. Opposite behaviour of the same posterior depending on use case is direct evidence that subjective-NLP uncertainty evaluation cannot be closed along a single axis.

\subsection{Selective prediction and misclassification detection}\label{sec:c5_main}

\begin{figure}[t]
\centering
\includegraphics[width=0.95\columnwidth]{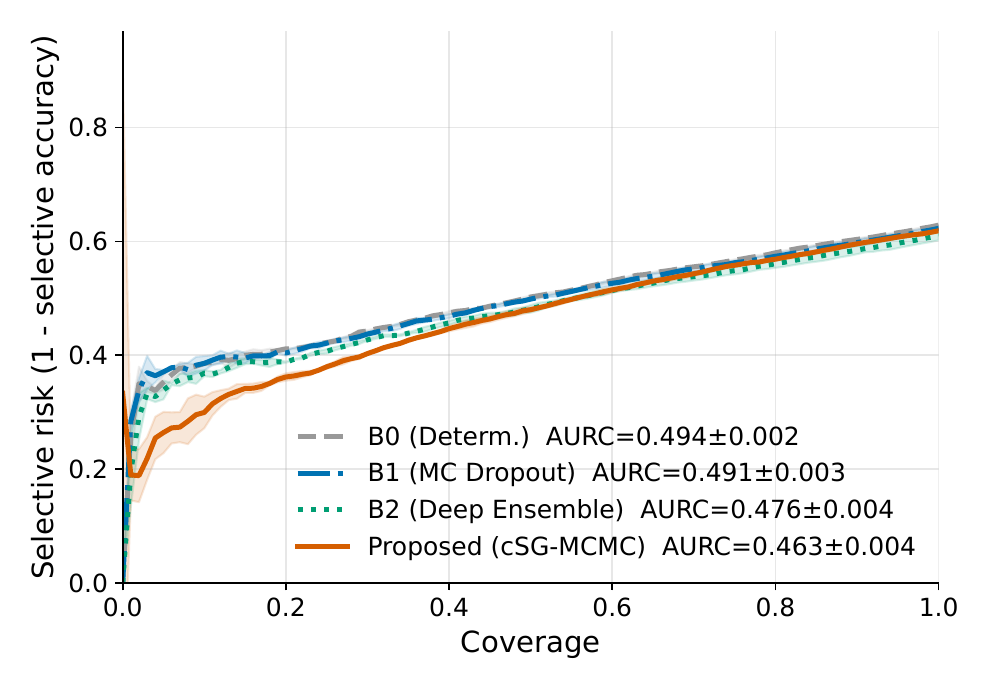}
\caption{Risk-coverage curves for selective prediction.}
\label{fig:c5_aurc}
\end{figure}

Table~\ref{tab:c5_aurc_auroc} shows selective-prediction metrics on test, soft label. With total entropy as rejector, the proposed method attains AURC $0.463 \pm 0.004$, the lowest among all baselines (B0 $0.494$, B1 $0.491$, B2 $0.476$), and AUROC $0.698 \pm 0.003$, the highest (baselines $0.679$--$0.683$); see Figure~\ref{fig:c5_aurc}. Even under the epistemic rejector, the proposed AUROC $0.639$ significantly surpasses B1 ($0.537$) and B2 ($0.526$).

That total entropy attains a higher AUROC than the epistemic score demonstrates the effectiveness of aggregating total uncertainty for reject-and-defer operation. Summing aleatoric and epistemic into a single metric surpasses baselines on the concrete downstream task of misclassification detection.

\begin{table*}[t]
\small
\centering
\setlength{\tabcolsep}{4pt}
\caption{Selective-prediction AURC and misclassification-detection AUROC.}
\label{tab:c5_aurc_auroc}
\begin{tabular}{l cccc}
\toprule
& \multicolumn{2}{c}{rejector = total entropy} & \multicolumn{2}{c}{rejector = epistemic} \\
\cmidrule(lr){2-3} \cmidrule(lr){4-5}
Method & AURC & AUROC & AURC & AUROC \\
\midrule
B0 & $0.494 \pm 0.002$ & $0.683 \pm 0.002$ & --- & --- \\
B1 & $0.491 \pm 0.003$ & $0.679 \pm 0.006$ & $0.585 \pm 0.005$ & $0.537 \pm 0.010$ \\
B2 & $0.476 \pm 0.004$ & $0.679 \pm 0.007$ & $0.585 \pm 0.014$ & $0.526 \pm 0.006$ \\
proposed & $0.463 \pm 0.004$ & $0.698 \pm 0.003$ & $0.491 \pm 0.015$ & $0.639 \pm 0.023$ \\
\bottomrule
\end{tabular}
\end{table*}

\subsection{Robustness ablation}\label{sec:results_robust}

Starting from canonical $\Ncyc = 8$, $T = 1.0$, $S = 5$, we independently vary $T$, $\Ncyc$, $S$ and evaluate sensitivity via mean total entropy (Table~\ref{tab:robust_summary}). For $T \in \{0.5, 1.0, 1.5\}$ the effect stays within a trivial range and is statistically indistinguishable; for $\Ncyc \in \{4, 8, 12\}$ we observe a LARGE effect with monotonic decrease (entropy $\downarrow$ as $\Ncyc \uparrow$). For $S \in \{3, 10, 20\}$ entropy increases monotonically but partially saturates at $S = 30$; this non-monotonicity limitation is discussed in \S\ref{sec:limitations}. All main claims are established under the canonical setting.

The contrast between the LARGE effect of $\Ncyc$ and the insensitivity of $T$ reveals a qualitative distinction between a structural hyperparameter directly determining posterior expressive power (cycles) and an auxiliary hyperparameter robust to fine-grained choices (temperature). For deployment, temperature tuning is light while the cycle count requires careful honest framing.

\begin{table}[t]
\footnotesize
\centering
\setlength{\tabcolsep}{2pt}
\caption{Robustness-ablation summary ($n=3$ seeds per level; 3 levels per axis; 4-level $S$ supplementary in App.~\ref{app:robust_full}).}
\label{tab:robust_summary}
\begin{tabular}{@{}l c c@{}}
\toprule
Ablation & ANOVA $F$ ($p$) & part. $\eta^2$ \\
\midrule
$\Ncyc \in \{4, 8, 12\}$ & $48.02$ ($2.0{\times}10^{-4}$) & $0.941$ \\
$T \in \{0.5, 1, 1.5\}$ & $0.20$ ($0.83$) & $0.061$ \\
$S \in \{3, 10, 20\}$ & $24.92$ ($1.2{\times}10^{-3}$) & $0.893$ \\
\bottomrule
\end{tabular}
\end{table}

\section{Discussion}\label{sec:discussion}

\begin{table*}[t!]
\footnotesize
\centering
\setlength{\tabcolsep}{4pt}
\caption{Comparison before and after applying post-hoc temperature scaling.}
\label{tab:t_scaling}
\begin{tabular}{llcccc}
\toprule
Method & Label & $T_{\text{opt}}$ & Brier pre/post & ECE pre/post & JSD pre/post \\
\midrule
B0 & hard & $0.787$ & $0.772 \to 0.767$ & $0.061 \to 0.068$ & --- \\
B0 & soft & $0.703$ & $0.788 \to 0.776$ & $0.085 \to 0.095$ & $0.534 \to 0.482$ \\
B1 & soft & $0.678$ & $0.790 \to 0.771$ & $0.104 \to 0.085$ & $0.537 \to 0.479$ \\
B2 & soft & $0.686$ & $0.779 \to 0.760$ & $0.107 \to 0.078$ & $0.527 \to 0.470$ \\
proposed & hard & $2.752$ & $0.847 \to 0.784$ & $0.248 \to 0.094$ & --- \\
proposed & soft & $2.015$ & $0.826 \to 0.788$ & $0.193 \to 0.079$ & $0.490 \to 0.528$ \\
\bottomrule
\end{tabular}
\end{table*}

\subsection{Posterior as an integrated uncertainty signal}\label{sec:disc_unification}

The central finding is that combining the cSG-MCMC posterior with soft-label learning simultaneously outperforms baselines along three axes: annotator-distribution fidelity (C2), per-emotion interpretability (C3), and misclassification detection in selective prediction (C5). These axes measure independent properties---geometric agreement of distributions, category-level detection of task-intrinsic ambiguity, and downstream reject-and-defer performance---each conventionally addressed by a separate method. Our results show they can be jointly attained from a single posterior, supporting an integrated-signal direction for subjective-NLP uncertainty evaluation rather than locally optimising individual metrics. Simultaneous improvement of all three axes from the same posterior suggests the advantage stems from a structural property of the posterior, not from overfitting to one metric.

\subsection{Independence of accuracy-centric and distribution-centric calibration}\label{sec:disc_two_dimensions}

C1 showed underperformance of the proposed method on hard-label argmax calibration while C2 showed strict dominance in annotator JSD: the same predictor yields opposite rankings. To make this explicit, we apply post-hoc temperature scaling \citep{guo2017oncalibration} to the $24$ runs (Table~\ref{tab:t_scaling}). For the proposed method, posterior cooling with $T_{\text{opt}} = 2.015$ markedly restores accuracy-centric calibration (Brier $0.826 \to 0.788$, ECE $0.193 \to 0.079$) but worsens annotator JSD ($0.490 \to 0.528$). The soft baselines instead move both metrics in the same direction (e.g.\ B2 soft, JSD $0.527 \to 0.470$, ECE $0.107 \to 0.078$). For B0 hard, ECE slightly worsens ($0.061 \to 0.068$): $T_{\text{opt}} = 0.787 < 1$ over-corrects via sharpening, a known sharpness-dependent phenomenon.

This bidirectional trade-off is empirical evidence that accuracy-centric calibration (argmax concentration) and distribution-centric calibration (annotator alignment) are independent dimensions: reporting only one renders the other invisible. We propose joint reporting of both, with explicit trade-off disclosure, as an honest reporting protocol. Since baselines lacking a distribution-centric axis move both metrics in the same direction under post-hoc calibration, the two-dimensional structure becomes visible only through the proposed method's posterior.

\subsection{Design implications and open questions}\label{sec:disc_design_implications}

Two design implications follow for deployment. First, $\Ncyc$ is a structural hyperparameter directly determining posterior expressivity, and its choice has a large impact (\S\ref{sec:results_robust}). All main results are established under canonical $\Ncyc = 8$, but varying it alters the expressivity / sampling-efficiency trade-off; this dependence must be honestly disclosed. Sensitivity to $T$, by contrast, lies in a negligible range and tuning effort is effectively zero. Second, the collapse of BALD acquisition in C4 shows that naive mutual-information acquisition without batch-level diversity fails structurally in long-tail subjective NLP. This is not merely negative but a positive contribution that pinpoints a task-specific failure mode. Solutions via class-balanced BatchBALD \citep{kirsch2019batchbald} or hybrid methods with minority-vote smoothing remain open.

\section{Conclusion}\label{sec:conclusion}

We proposed a predictor combining a cSG-MCMC head on a frozen RoBERTa backbone with soft-label cross-entropy training, and a five-axis framework (C1--C5) for uncertainty quality in subjective NLP. On the $28$-emotion GoEmotions task, the proposed method consistently improves over baselines in annotator-distribution JSD and selective-prediction AURC / AUROC, and its per-emotion aleatoric uncertainty correlates with the annotator disagreement rate. The bidirectional effect of post-hoc temperature scaling provides empirical evidence that accuracy-centric and distribution-centric calibration are independent dimensions, motivating an honest joint-reporting protocol. Sensitivity along $\Ncyc$ is honestly reported and all main claims are positioned under the canonical setting.

Future work includes generalisation to other subjective NLP tasks (EmoBank, SemEval-2018 affect) and non-English corpora, and extending the Bayesian treatment from the linear head to full-network MCMC.

\section*{Limitations}\label{sec:limitations}

This work has several limitations. First, validation is restricted to a single English dataset, GoEmotions. Generalisation to other subjective NLP tasks in which annotator disagreement is an intrinsic structure---such as EmoBank \citep{buechel2017emobank}, the multi-annotator empathy/distress corpus \citep{buechel2018modeling}, and the SemEval-2018 Task 1 affect corpus \citep{mohammad2018semeval}---and to non-English corpora is anticipated within this framework but has not been empirically verified.

The Bayesian treatment is limited to the linear head only. We freeze RoBERTa and apply cSG-MCMC only to the linear head ($21{,}532$ trainable parameters out of approximately $125$M). This is a design that separates the comparison between pre-training quality and the uncertainty mechanism and that suppresses computational cost, but it intentionally restricts the expressive power of the posterior. Extension to full-network MCMC remains an unexplored direction.

With respect to hyperparameter sensitivity, the proposed method is not fully robust. Along the $\Ncyc$ axis we observe a monotonic decrease and a LARGE effect (partial $\eta^2 = 0.941$, Cohen's $d_{12-4} = -8.26$), so the proposed method is structurally sensitive to the choice of the number of cycles. We make explicit that all main claims are established under the canonical $\Ncyc = 8$ setting. For samples per cycle $S \in \{3, 10, 20\}$, mean total entropy increases monotonically, whereas the post-hoc additional $S = 30$ shows a slight saturation; the monotonic increase therefore holds only within the canonical range.

The per-category Spearman $\rho$ in \S\ref{sec:c3_main} is computed over only $K = 23$--$24$ categories, and the bootstrap $95\%$ CIs are wide on the validation split ($[0.10, 0.78]$ under soft and $[0.14, 0.84]$ under hard). The decrease from validation to test (soft: $0.489 \to 0.336$; hard: $0.556 \to 0.417$) is consistent with the range of small-$K$ sampling variability rather than performance degradation, but strong claims about the absolute magnitude of the correlation are difficult given our statistical power.

The BALD failure mode reported in \S\ref{sec:results_c4} ($27$-class GoEmotions setup; final-iteration validation accuracy of BALD $\{0.297, 0.084, 0.201\}$ vs entropy / random $\approx 0.46$) is a structural failure intrinsic to long-tail subjective NLP, and validation of class-balanced BatchBALD variants \citep{kirsch2019batchbald} or hybrid methods that incorporate minority-vote smoothing is left for future work. Our uncertainty score is effective as a rejector (C5), but exhibits a confirmed asymmetry in that it does not function as an acquisition signal within the BALD framework.

\section*{Ethics Statement}

This work uses the publicly available GoEmotions corpus \citep{demszky2020}; no new data collection or human-subject experiments were conducted. GoEmotions is built from public Reddit comments and was anonymised by its providers, but text originating from social media may contain cultural, geographic, and demographic biases. When deploying models trained in this work in real-world settings, it is necessary to be aware that such biases may be reflected in emotion predictions.

The main contribution of this work is an evaluation framework and a predictor that preserve annotator disagreement as a structure rather than discarding it. This supports a direction in subjective NLP that does not render minority perspectives invisible through majority voting, and carries ethical implications for applications that respect the diversity of human interpretation---such as content moderation and mental-health-related tasks. At the same time, the category-level interpretability of aleatoric uncertainty and the effectiveness of reject-and-defer operation discussed in this work also have an aspect that may promote the automation of machine judgement; the question of which decisions to escalate to a human in a human-in-the-loop design therefore continues to require careful deliberation.

In terms of computational cost, all experiments in this work can be executed on a single GPU, and the wall-clock training time of the proposed method is about $0.89$ times that of the B0 baseline. Because large-scale pre-training is not required, the environmental footprint of this work is limited.

\section*{Use of AI Assistants}

AI assistants were used in two capacities. First, for coding assistance (e.g., code refactoring and debugging support) during the implementation of our experiments. Second, the manuscript was originally written in Japanese and translated into English using an AI assistant; the authors subsequently performed manual post-editing to verify factual accuracy and to refine the English expression. All research questions, experimental design, analyses, interpretations, and scientific claims are entirely the authors' own, and the authors take full responsibility for the final content.

\bibliography{refs}

@inproceedings{demszky2020,
  author    = {Demszky, Dorottya and Movshovitz-Attias, Dana and Ko, Jeongwoo and Cowen, Alan and Nemade, Gaurav and Ravi, Sujith},
  title     = {{GoEmotions}: A Dataset of Fine-Grained Emotions},
  booktitle = {Proceedings of the 58th Annual Meeting of the Association for Computational Linguistics},
  year      = {2020},
  pages     = {4040--4054},
  doi = {10.18653/v1/2020.acl-main.372},
}

@inproceedings{pavlick2019,
  author    = {Pavlick, Ellie and Kwiatkowski, Tom},
  title     = {Inherent Disagreements in Human Textual Inferences},
  booktitle = {Transactions of the Association for Computational Linguistics},
  volume    = {7},
  year      = {2019},
  pages     = {677--694},
  doi = {10.1162/tacl_a_00293}
}

@inproceedings{plank2022,
  author    = {Plank, Barbara},
  title     = {The "Problem" of Human Label Variation: On Ground Truth in Data, Modeling and Evaluation},
  booktitle = {Proceedings of the 2022 Conference on Empirical Methods in Natural Language Processing},
  year      = {2022},
  pages     = {10671--10682},
  doi = {10.18653/v1/2022.emnlp-main.731}
}

@article{uma2021,
  author    = {Uma, Alexandra N. and Fornaciari, Tommaso and Hovy, Dirk and Paun, Silviu and Plank, Barbara and Poesio, Massimo},
  title     = {Learning from Disagreement: A Survey},
  journal   = {Journal of Artificial Intelligence Research},
  volume    = {72},
  year      = {2021},
  pages     = {1385--1470},
  doi = {10.1613/jair.1.12752}
}

@inproceedings{welling2011sgld,
  author    = {Welling, Max and Teh, Yee Whye},
  title     = {Bayesian Learning via Stochastic Gradient Langevin Dynamics},
  booktitle = {Proceedings of the 28th International Conference on Machine Learning},
  year      = {2011},
  pages     = {681--688},
  doi = {10.5555/3104482.3104568}
}

@inproceedings{chen2014sgmcmc,
  author    = {Chen, Tianqi and Fox, Emily B. and Guestrin, Carlos},
  title     = {Stochastic Gradient {H}amiltonian {M}onte {C}arlo},
  booktitle = {Proceedings of the 31st International Conference on Machine Learning},
  year      = {2014},
  pages     = {1683--1691},
  doi = {10.5555/3044805.3045080}
}

@inproceedings{zhang2020csgmcmc,
  author    = {Zhang, Ruqi and Li, Chunyuan and Zhang, Jianyi and Chen, Changyou and Wilson, Andrew Gordon},
  title     = {Cyclical Stochastic Gradient {MCMC} for {B}ayesian Deep Learning},
  booktitle = {International Conference on Learning Representations},
  year      = {2020},
}

@inproceedings{depeweg2018,
  author    = {Depeweg, Stefan and Hernandez-Lobato, Jose Miguel and Doshi-Velez, Finale and Udluft, Steffen},
  title     = {Decomposition of Uncertainty in Bayesian Deep Learning for Efficient and Risk-sensitive Learning},
  booktitle = {Proceedings of the 35th International Conference on Machine Learning},
  year      = {2018},
  pages     = {1192--1201},
}

@inproceedings{kendall2017,
  author    = {Kendall, Alex and Gal, Yarin},
  title     = {What Uncertainties Do We Need in {B}ayesian Deep Learning for Computer Vision?},
  booktitle = {Proceedings of the 31st International Conference on Neural Information Processing Systems},
  year      = {2017},
  pages     = {5580--5590},
}

@article{houlsby2011bald,
  author    = {Houlsby, Neil and Husz{\'a}r, Ferenc and Ghahramani, Zoubin and Lengyel, M{\'a}t{\'e}},
  title     = {{B}ayesian Active Learning for Classification and Preference Learning},
  journal = {arXiv},
  year      = {2011},
  doi = {10.48550/arXiv.1112.5745}
}

@inproceedings{kirsch2019batchbald,
  author    = {Kirsch, Andreas and van Amersfoort, Joost and Gal, Yarin},
  title     = {{B}atch{BALD}: Efficient and Diverse Batch Acquisition for Deep {B}ayesian Active Learning},
  booktitle = {Proceedings of the 33rd International Conference on Neural Information Processing Systems},
  year      = {2019},
  pages     = {1--12},
}

@inproceedings{wenzel2020,
  author    = {Wenzel, Florian and Roth, Kevin and Veeling, Bastiaan and Swiatkowski, Jakub and Tran, Linh and Mandt, Stephan and Snoek, Jasper and Salimans, Tim and Jenatton, Rodolphe and Nowozin, Sebastian},
  title     = {How Good is the {B}ayes Posterior in Deep Neural Networks Really?},
  booktitle = {Proceedings of the 37th International Conference on Machine Learning},
  year      = {2020},
  number = {949},
  pages = {10248--10259},
  doi = {10.5555/3524938.3525887}
}

@inproceedings{aitchison2021,
  author    = {Aitchison, Laurence},
  title     = {A Statistical Theory of Cold Posteriors in Deep Neural Networks},
  booktitle = {International Conference on Learning Representations},
  year      = {2021},
  doi = {10.48550/arXiv.2008.05912}
}

@article{liu2019roberta,
  author    = {Liu, Yinhan and Ott, Myle and Goyal, Naman and Du, Jingfei and Joshi, Mandar and Chen, Danqi and Levy, Omer and Lewis, Mike and Zettlemoyer, Luke and Stoyanov, Veselin},
  title     = {{R}o{BERT}a: A Robustly Optimized {BERT} Pretraining Approach},
  journal   = {arXiv},
  year      = {2019},
  doi = {10.48550/arXiv.1907.11692}
}

@inproceedings{gal2016dropout,
  author    = {Gal, Yarin and Ghahramani, Zoubin},
  title     = {Dropout as a {B}ayesian Approximation: Representing Model Uncertainty in Deep Learning},
  booktitle = {Proceedings of the 33rd International Conference on Machine Learning},
  year      = {2016},
  volume = {48},
  pages     = {1050--1059},
  doi = {10.5555/3045390.3045502}
}

@inproceedings{lakshminarayanan2017,
  author    = {Lakshminarayanan, Balaji and Pritzel, Alexander and Blundell, Charles},
  title     = {Simple and Scalable Predictive Uncertainty Estimation using Deep Ensembles},
  booktitle = {Proceedings of the 31st International Conference on Neural Information Processing Systems},
  year      = {2017},
  pages     = {6405--6416},
  doi = {10.5555/3295222.3295387}
}

@inproceedings{guo2017oncalibration,
  author    = {Guo, Chuan and Pleiss, Geoff and Sun, Yu and Weinberger, Kilian Q.},
  title     = {On Calibration of Modern Neural Networks},
  booktitle = {Proceedings of the 34th International Conference on Machine Learning},
  year      = {2017},
  pages     = {1321--1330},
  doi = {10.5555/3305381.3305518}
}

@inproceedings{geifman2017selective,
  author    = {Geifman, Yonatan and El-Yaniv, Ran},
  title     = {Selective Classification for Deep Neural Networks},
  booktitle = {Proceedings of the 31st International Conference on Neural Information Processing Systems},
  year      = {2017},
  pages     = {4885--4894},
}

@inproceedings{peterson2019humanuncertainty,
  author    = {Peterson, Joshua and Battleday, Ruairidh and Griffiths, Thomas and Russakovsky, Olga},
  title     = {Human Uncertainty Makes Classification More Robust},
  booktitle = {Proceedings of the 2019 IEEE/CVF International Conference on Computer Vision},
  year      = {2019},
  pages     = {9616--9625},
  doi = {10.1109/ICCV.2019.00971}
}

@article{hinton2015distilling,
  author    = {Hinton, Geoffrey and Vinyals, Oriol and Dean, Jeff},
  title     = {Distilling the Knowledge in a Neural Network},
  journal   = {NIPS Deep Learning and Representation Learning Workshop},
  year      = {2015},
  doi = {10.48550/arXiv.1503.02531}
}

@inproceedings{kumar2022finetuning,
  author    = {Kumar, Ananya and Raghunathan, Aditi and Jones, Robbie and Ma, Tengyu and Liang, Percy},
  title     = {Fine-Tuning Can Distort Pretrained Features and Underperform Out-of-Distribution},
  booktitle = {International Conference on Learning Representations},
  year      = {2022},
  doi = {10.48550/arXiv.2202.10054}
}

@article{buechel2017emobank,
  author    = {Buechel, Sven and Hahn, Udo},
  title     = {{E}mo{B}ank: Studying the Impact of Annotation Perspective and Representation Format on Dimensional Emotion Analysis},
  journal   = {Proceedings of the 15th Conference of the European Chapter of the Association for Computational Linguistics},
  pages     = {578--585},
  year      = {2017}
}

@inproceedings{buechel2018modeling,
  author    = {Buechel, Sven and Buffone, Anneke and Slaff, Barry and Ungar, Lyle and Sedoc, Jo{\~a}o},
  title     = {Modeling Empathy and Distress in Reaction to News Stories},
  booktitle = {Proceedings of the 2018 Conference on Empirical Methods in Natural Language Processing},
  pages     = {4758--4765},
  year      = {2018},
  doi = {10.18653/v1/D18-1507}
}

@inproceedings{mohammad2018semeval,
  author    = {Mohammad, Saif and Bravo-Marquez, Felipe and Salameh, Mohammad and Kiritchenko, Svetlana},
  title     = {{S}em{E}val-2018 Task 1: Affect in Tweets},
  booktitle = {Proceedings of The 12th International Workshop on Semantic Evaluation},
  pages     = {1--17},
  year      = {2018},
  doi = {10.18653/v1/S18-1001}
}

@article{holm1979simple,
  author    = {Holm, Sture},
  title     = {A Simple Sequentially Rejective Multiple Test Procedure},
  journal   = {Scandinavian Journal of Statistics},
  volume    = {6},
  number    = {2},
  pages     = {65--70},
  year      = {1979},
}

@inproceedings{hendrycks2017baseline,
  author    = {Hendrycks, Dan and Gimpel, Kevin},
  title     = {A Baseline for Detecting Misclassified and Out-of-Distribution Examples in Neural Networks},
  booktitle = {Proceedings of the 5th International Conference on Learning Representations},
  year      = {2017},
  doi = {10.48550/arXiv.1610.02136}
}

@article{baan2022stop,
  author    = {Baan, Joris and Aziz, Wilker and Plank, Barbara and Fernandez, Raquel},
  title     = {Stop Measuring Calibration When Humans Disagree},
  journal   = {Proceedings of the 2022 Conference on Empirical Methods in Natural Language Processing},
  year      = {2022},
  pages = {1892-1915},
  doi = {10.18653/v1/2022.emnlp-main.124}
}

@article{frenda2024perspectivist,
  author    = {Frenda, Simona and Abercrombie, Gavin and Basile, Valerio and Pedrani, Alessandro and Panizzon, Raffaella and Cignarella, Alessandra Teresa and Marco, Cristina and Bernardi, Davide},
  title     = {Perspectivist Approaches to Natural Language Processing: A Survey},
  journal   = {Language Resources and Evaluation},
  volume    = {59},
  year      = {2025},
  pages = {1719-1746},
  doi       = {10.1007/s10579-024-09766-4},
}

@inproceedings{baan2024interpreting,
  author    = {Baan, Joris and Fern{\'a}ndez, Raquel and Plank, Barbara and Aziz, Wilker},
  title     = {Interpreting Predictive Probabilities: Model Confidence or Human Label Variation?},
  booktitle = {Proceedings of the 18th Conference of the European Chapter of the Association for Computational Linguistics},
  pages     = {268--277},
  year      = {2024},
  doi      = {10.18653/v1/2024.eacl-short.24},
}

@inproceedings{vandermeer2024annotator,
  author    = {van der Meer, Michiel and Falk, Neele and Murukannaiah, Pradeep K. and Liscio, Enrico},
  title     = {Annotator-Centric Active Learning for Subjective {NLP} Tasks},
  booktitle = {Proceedings of the 2024 Conference on Empirical Methods in Natural Language Processing},
  pages     = {18537--18555},
  year      = {2024},
  doi      = {10.18653/v1/2024.emnlp-main.1031},
}

@inproceedings{sale2024labelwise,
  author    = {Sale, Yusuf and Hofman, Paul and L{\"o}hr, Timo and Wimmer, Lisa and Nagler, Thomas and H{\"u}llermeier, Eyke},
  title     = {Label-wise Aleatoric and Epistemic Uncertainty Quantification},
  booktitle = {Proceedings of the Fortieth Conference on Uncertainty in Artificial Intelligence},
  series    = {Proceedings of Machine Learning Research},
  volume    = {147},
  year      = {2024},
  pages = {3159--3179},
  doi      = {10.5555/3702676.3702823},
}

@inproceedings{wimmer2023quantifying,
  author    = {Wimmer, Lisa and Sale, Yusuf and Hofman, Paul and Bischl, Bernd and H{\"u}llermeier, Eyke},
  title     = {Quantifying Aleatoric and Epistemic Uncertainty in Machine Learning: Are Conditional Entropy and Mutual Information Appropriate Measures?},
  booktitle = {Proceedings of the Thirty-Ninth Conference on Uncertainty in Artificial Intelligence (UAI)},
  series    = {Proceedings of Machine Learning Research},
  volume    = {216},
  year      = {2023},
  number = {213},
  pages = {2282--2292},
  doi = {10.5555/3625834.3626047}
}

@article{brier1950verification,
  author  = {Brier, Glenn W.},
  title   = {Verification of Forecasts Expressed in Terms of Probability},
  journal = {Monthly Weather Review},
  volume  = {78},
  number  = {1},
  pages   = {1--3},
  year    = {1950},
  doi     = {10.1175/1520-0493(1950)078<0001:VOFEIT>2.0.CO;2},
}

@article{keito2026bridge,
  author       = {Inoshita, Keito},
  title        = {Bridging the Silos in Affective {AI}: A Critical Perspective from Data to Society},
  year         = {2026},
  journal    = {SSRN},
}

\appendix
\onecolumn

\section*{Appendix}

\section{Implementation details and software environment}\label{app:impl_env}\label{app:method_details}\label{app:setup_details}

Regarding the implementation of the frozen backbone, we set \texttt{requires\_grad=False} on all parameters of RoBERTa (embeddings, the $12$ encoder layers, the pooler, and the weights and biases of LayerNorm), and ensure at the implementation level that gradients do not propagate to the backbone during the backward pass. Only the head ($21{,}532$ trainable parameters) is therefore subject to training. Because B2 (Deep Ensemble) uses $5$ independent heads, it has $107{,}660$ trainable parameters in total. Inputs are tokenised at a maximum length of $128$ tokens (more than $99.9\%$ of comments fit without truncation), and the mean-pooled $[\mathrm{CLS}]$ representation is fed into a linear head ($768 \to 28$). Forward and backward passes are executed in bfloat16 mixed precision, while loss accumulation, gradient computation, and the noise injection of cSG-MCMC are performed in float32 for numerical stability. Gradient clipping is applied with $\|g\|_2 \le 1.0$ for B0 / B1 / B2 and $\|g\|_2 \le 5.0$ for the proposed method. The relaxed clipping threshold for the proposed method is designed to accommodate updates that include Gaussian-noise injection during the sampling phase.

The soft-label loss $\kl(q_i \| p_\theta(\cdot|x_i))$ is computed with an implementation equivalent to PyTorch's \texttt{F.cross\_entropy(logits, q, reduction='mean')} (the soft-target mode that is invoked when the target is a probability distribution).

The software environment uses Python $3.10$, PyTorch $2.5.1$ (CUDA $12.1$ build), HuggingFace transformers $4.52.4$, NumPy $1.26$, SciPy $1.13$, scikit-learn $1.5$, statsmodels $0.14$, and matplotlib $3.8$. To ensure determinism, \texttt{torch.use\_deterministic\_algorithms(True)} is enabled, and seeds $\{42, 43, 44\}$ control PyTorch, NumPy, the random module, and the HuggingFace data loader in a unified manner. GoEmotions \citep{demszky2020} is released under Apache 2.0 and RoBERTa-base \citep{liu2019roberta} under the MIT license, and our use (frozen backbone, research use) is consistent with the intended use of both.

\section{Statistical protocol details}\label{app:stats}

We supplement here the implementation details of the testing plan listed in \S\ref{sec:metrics} of the main body. The two-way ANOVA on \emph{method} and \emph{label} with Type II SS is executed with \texttt{statsmodels} v$0.14$ and cross-validated with an in-house closed-form balanced ANOVA. For the main grid $N = 24$, $\mathrm{df}_{\text{method}}=3$, $\mathrm{df}_{\text{label}}=1$, $\mathrm{df}_{\text{method}\times\text{label}}=3$, $\mathrm{df}_{\text{residual}}=16$, and the SS / F / $p$ from statsmodels and the closed-form computation agree to all displayed digits.

Paired $t$-tests are executed two-sided on the run-level means across $3$ seeds, and are sequentially corrected with Holm--Bonferroni \citep{holm1979simple} against the $3$ baselines (family-wise $\alpha = 0.05$, step-down sequential rejection). Bootstrap $95\%$ CIs use $1{,}000$ iterations and the percentile method, and we also report Cohen's $d$ and partial $\eta^2$. The strict dominance defined in \S\ref{sec:metrics} of the main body requires both (D1) paired differences signed in the improvement direction across all seeds and (D2) Holm-corrected $p < 0.05$, and constitutes a strict criterion.

The per-category annotator disagreement rate $\delta^{(c)}$ is computed once from the $3$-rater vote of the original GoEmotions data and is identical across all runs. The per-category Spearman $\rho$ is computed for emotion categories with $30$ or more examples within a split ($K \approx 23$--$24$).

\section{Dataset details}\label{app:dataset_details}

We adopt the canonical splits of GoEmotions \citep{demszky2020}: train / validation / test $= 43{,}410 / 5{,}426 / 5{,}427$. Each example has a vote vector over $C = 28$ emotion categories from $A_i \in \{3, 4, 5\}$ raters. The annotator pool consists of $82$ US-based native English speakers, and the demographic distribution is disclosed by \citet{demszky2020}. Pre-processing is limited to the canonical pre-processing of GoEmotions (no lowercasing, special tokens preserved); no additional re-tokenisation or text editing is performed.

Per-category sample counts form a long-tail distribution. The top-$5$ frequencies in the train split are neutral ($14{,}219$), admiration ($4{,}130$), approval ($2{,}939$), gratitude ($2{,}662$), and annoyance ($2{,}470$); the bottom-$5$ frequencies are grief ($77$), pride ($111$), relief ($153$), nervousness ($164$), and embarrassment ($303$). This long-tail structure is the structural cause of the BALD failure mode in \S\ref{sec:results_c4}.

\section{C2: JSD extended results}\label{app:c2_full}

For the JSD between the predictive distribution and the empirical annotator distribution shown in \S\ref{sec:c2_main} of the main body, we provide the seed-mean details for the validation split, soft label. B0 soft is $0.535 \pm 0.002$, B1 soft is $0.538 \pm 0.003$, B2 soft is $0.528 \pm 0.003$, and proposed soft is $0.490 \pm 0.003$; the proposed method strictly dominates every baseline with Holm-corrected paired $t$, $p < 0.01$. Under hard labels as well, the proposed method ($0.491 \pm 0.002$) consistently undercuts B0 ($0.520$), B1 ($0.524$), and B2 ($0.516$), and the absolute difference between validation and test stays within $0.003$--$0.005$ bits. As auxiliary quantities, the mean KL divergence is $2.33$ for the proposed method vs $1.51$--$1.54$ for the baselines, and the total variation (TV) distance is $0.620$ for the proposed method vs $0.700$ for the baselines; both indicate the same ordering as JSD.

\section{C3: per-emotion $\rho$ extended results}\label{app:c3_full}

For the per-emotion Spearman $\rho$ in \S\ref{sec:c3_main} of the main body, we report values on the test split. They decay to $\rho = 0.336$ under soft and $\rho = 0.417$ under hard, but this is consistent with the range of small-$K$ sampling variability for $K \approx 23$--$24$, and the bootstrap CIs overlap between the validation and test splits. The qualitative ordering (proposed $>$ baselines) is preserved on both splits.

We show the per-emotion uncertainty heatmaps below. Category orderings are sorted within each label mode in descending order of per-category $\rho$, and the orderings under hard and soft do not match. The neutral category, whose frequency is so high that it complicates interpretation, is excluded, and the heatmaps are constructed over the remaining $27$ categories.

\begin{figure}[h]
\centering
\includegraphics[width=0.85\textwidth]{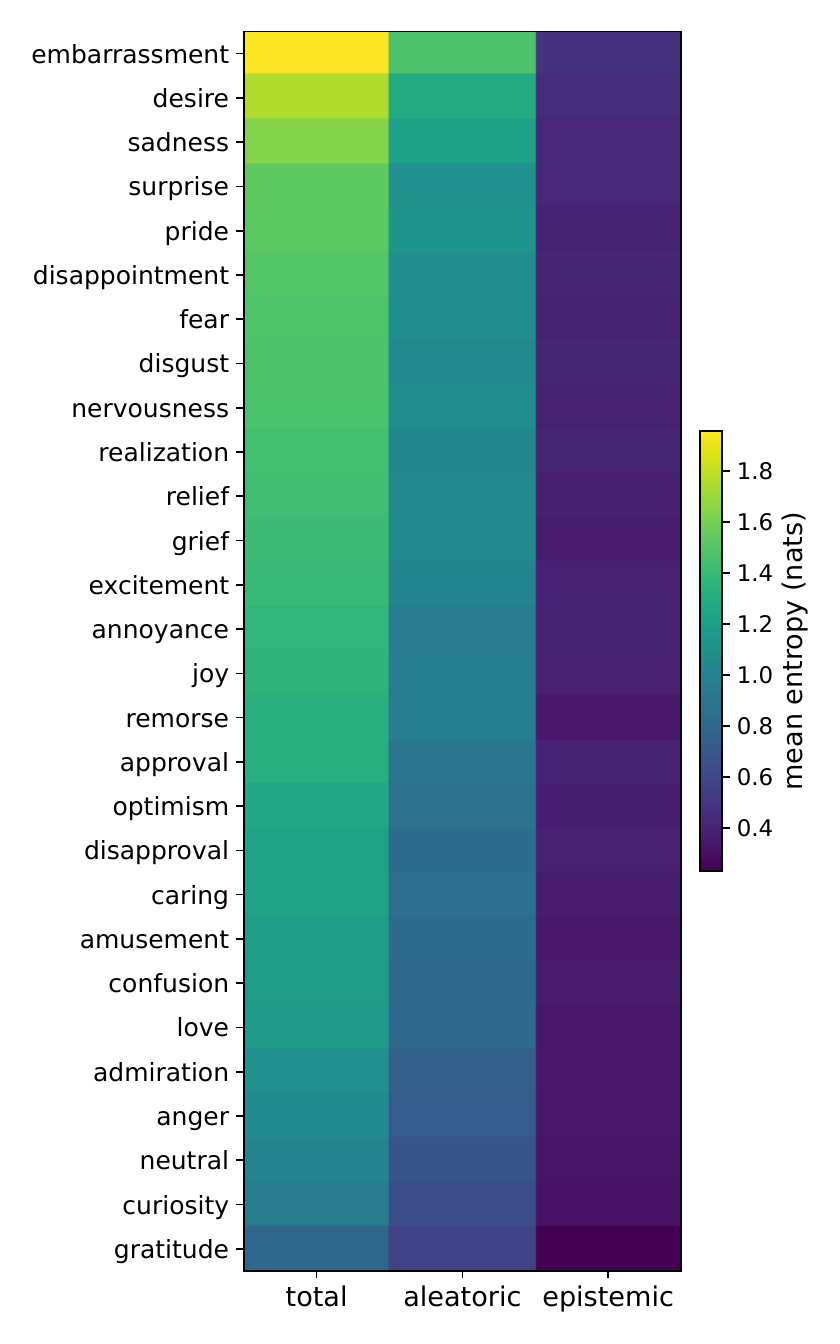}
\caption{Per-emotion uncertainty heatmap (hard label, test split).}
\label{fig:per_emotion_hard}
\end{figure}

\begin{figure}[h]
\centering
\includegraphics[width=0.85\textwidth]{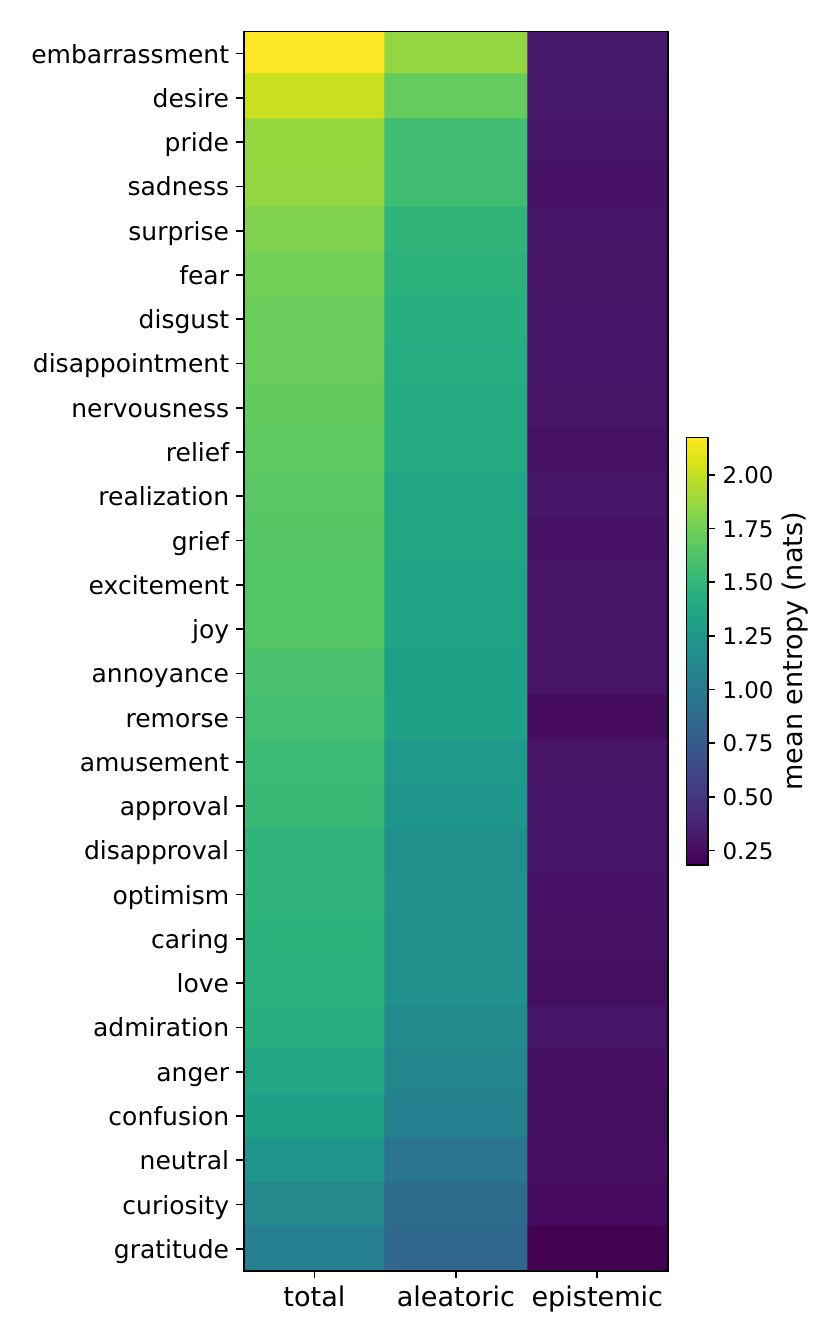}
\caption{Per-emotion uncertainty heatmap (soft label, test split).}
\label{fig:per_emotion_soft}
\end{figure}

\section{C5 AURC / AUROC details}\label{app:c5_full}

The values in Table~\ref{tab:c5_aurc_auroc} of the main body integrate the AURC / AUROC of all baselines and the proposed method on the test split, soft label. In an auxiliary analysis with the aleatoric rejector, the proposed AUROC is $0.698 \pm 0.002$, the highest among all methods, matching the value with the total-entropy rejector. This shows that aleatoric uncertainty alone is also effective for misclassification detection; however, as discussed in \S\ref{sec:c5_main} of the main body, the conclusion that total entropy (aleatoric $+$ epistemic) is the most informative rejector is unchanged. The base error rate is $0.61$--$0.62$ across all methods (with a difference of $\le 0.02$), so the AURC / AUROC differences arise from the quality of the rejector signal rather than from variation in base error.

\section{Robustness ablation extended evidence}\label{app:robust_full}

This section supplements the numerical basis for the three-axis ablation reported in Table~\ref{tab:robust_summary} of \S\ref{sec:results_robust} and the 4-level aggregation along the $S$ axis.

For the $\Ncyc$ ablation, we ran a total of $9$ runs ($3$ seeds each) at $\Ncyc \in \{4, 8, 12\}$. The mean total entropy decreases monotonically: $1.314 \pm 0.034$ nats at $\Ncyc = 4$, $1.164 \pm 0.037$ nats at $\Ncyc = 8$ (canonical), and $1.038 \pm 0.033$ nats at $\Ncyc = 12$. The pairwise effect size against the canonical setting is Cohen's $d = -8.26$ and mean difference $= -0.275$ [bootstrap $95\%$ CI: $-0.318, -0.232$] nats for $\Ncyc = 12$ vs $\Ncyc = 4$.

For the canonical $3$ levels of the $S$ ablation, we ran a total of $9$ runs at $S \in \{3, 10, 20\}$. The mean total entropy increases monotonically: $S_3 = 1.112 \pm 0.030$, $S_{10} = 1.204 \pm 0.015$, and $S_{20} = 1.222 \pm 0.011$ nats. For $S_{20}$ vs $S_3$, Cohen's $d = +4.85$ and mean difference $+0.110$ [$95\%$ CI: $+0.077, +0.135$] nats.

For honest reporting, we also report a 4-level aggregation that includes $S = 30$ ($3$ seeds). $S_{30} = 1.164 \pm 0.037$ nats, so the monotonic increase saturates at $S = 20$ and slightly decreases at $S = 30$. The 4-level ANOVA gives $F(3, 8) = 10.997$, $p = 3.28 \times 10^{-3}$, and partial $\eta^2 = 0.805$, so the effect size remains LARGE. The monotonic narrative in \S\ref{sec:results_robust} of the main body is therefore restricted to the canonical $\{3, 10, 20\}$ range.

For the $T$ ablation, we ran a total of $9$ runs at $T \in \{0.5, 1.0, 1.5\}$. The mean total entropy is $T_{0.5} = 1.180 \pm 0.040$, $T_{1.0} = 1.164 \pm 0.037$, and $T_{1.5} = 1.179 \pm 0.031$ nats (max Cohen's $d = 0.46$, max mean difference $0.016$ nats). All pairwise $95\%$ bootstrap CIs contain $0$, and neither direction of posterior tempering---cold or hot---rejects the null hypothesis against canonical $T = 1.0$. Under the power constraint of $n = 3$/level, we restrict the articulation to ``the observed effect lies within a trivial range''.

\begin{figure}[h]
\centering
\includegraphics[width=0.75\textwidth]{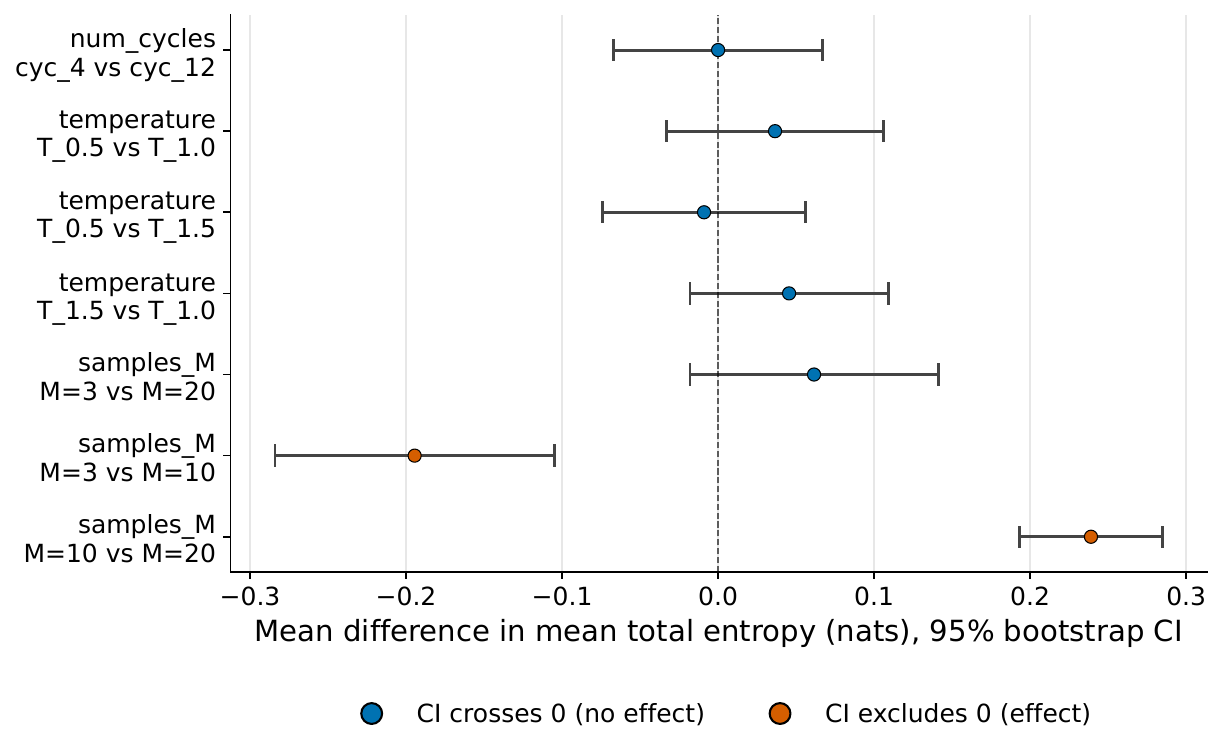}
\caption{Forest plot of the robustness ablation (pairwise mean differences with $95\%$ bootstrap CIs, in units of nats).}
\label{fig:effect_size_robust}
\end{figure}

\section{Post-hoc temperature scaling details}\label{app:tscaling_full}

This section supplements Table~\ref{tab:t_scaling} of \S\ref{sec:disc_two_dimensions} in the main body. For each of the $24$ main runs, we obtain a single scalar $T_{\text{opt}}$ by maximum-likelihood NLL on the validation set, and re-evaluate by dividing the test-split logits by $T_{\text{opt}}$ \citep{guo2017oncalibration}. For multi-member methods (B1 / B2 / proposed), $T_{\text{opt}}$ is applied to the per-member logits and the member average is then taken.

The per-run percentage reductions for the proposed method (soft) are Brier $-4.61\%$ ($0.8255 \to 0.7875$) and ECE $-58.90\%$ ($0.1931 \to 0.0794$), and for the proposed method (hard) ECE $-62.26\%$ ($0.2485 \to 0.0938$); these are consistent with the standard rounded values in the main body ($-4.6\%$ / $-58.9\%$ / $-62.3\%$).

Regarding rank invariance, T-scaling is exactly order-preserving for single-head methods (B0), but for multi-member methods the argmax slightly permutes after the member average over per-member-applied scaling. Across the $24$ runs we measure Spearman $\rho > 0.97$ for top-$1$ confidence and Spearman $\rho > 0.94$ for entropy, confirming that rank-based metrics (such as AURC) are essentially invariant.

\begin{figure}[h]
\centering
\includegraphics[width=0.4\textwidth]{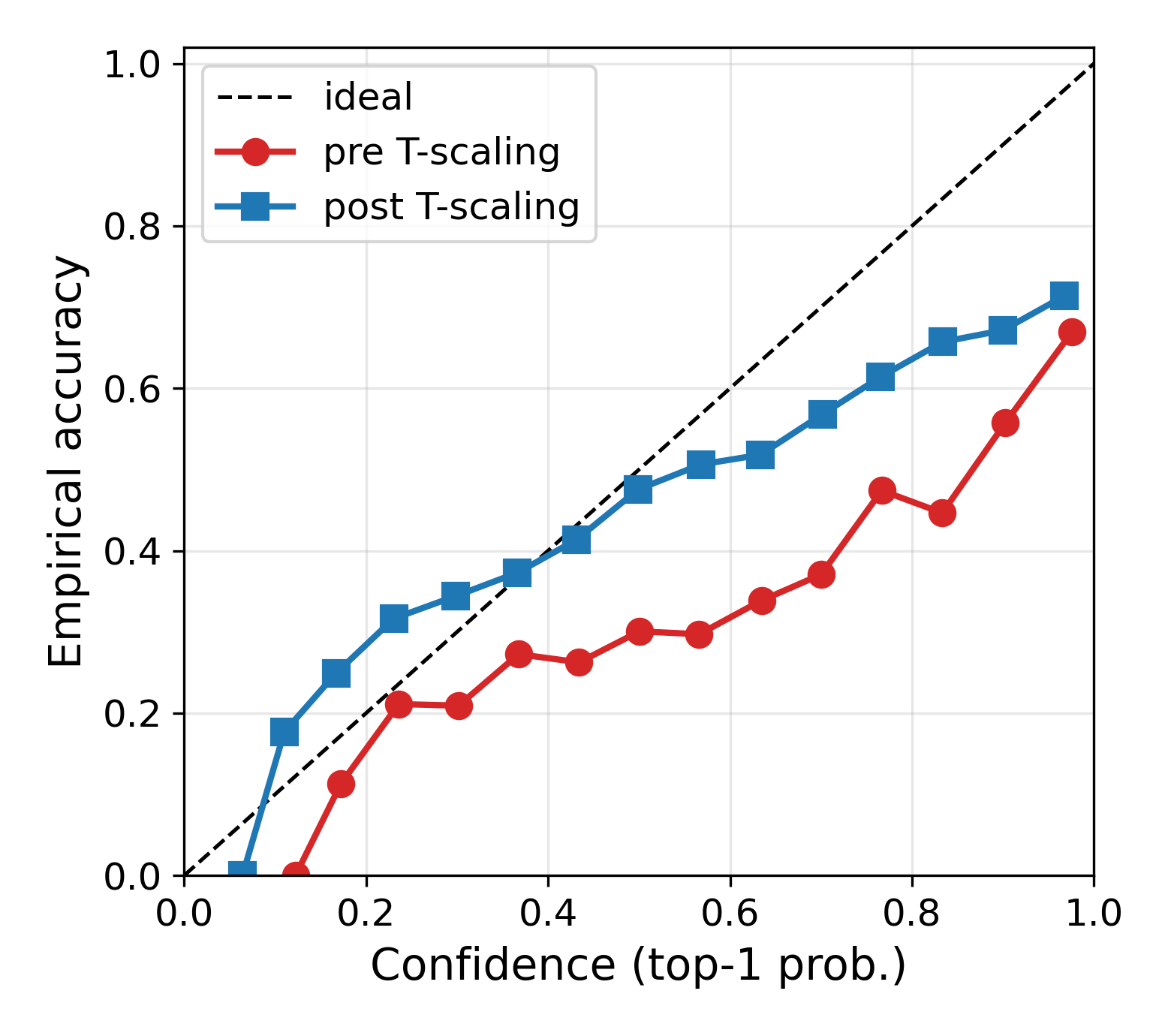}
\hfill
\includegraphics[width=0.4\textwidth]{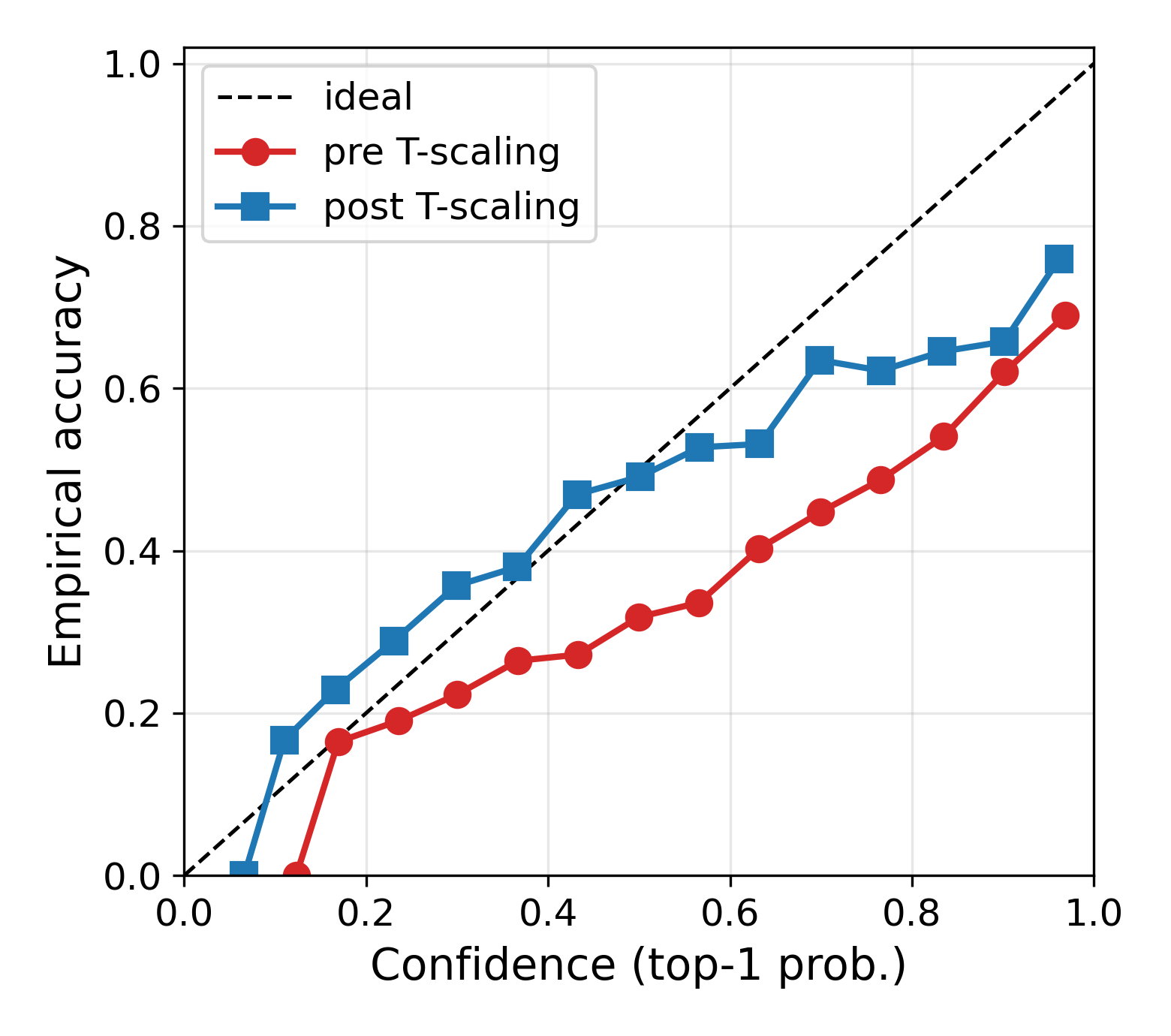}
\caption{Reliability diagrams for the proposed method (left: hard label, right: soft label; test split).}
\label{fig:reliability_proposed}
\end{figure}

\begin{figure}[h]
\centering
\includegraphics[width=0.24\textwidth]{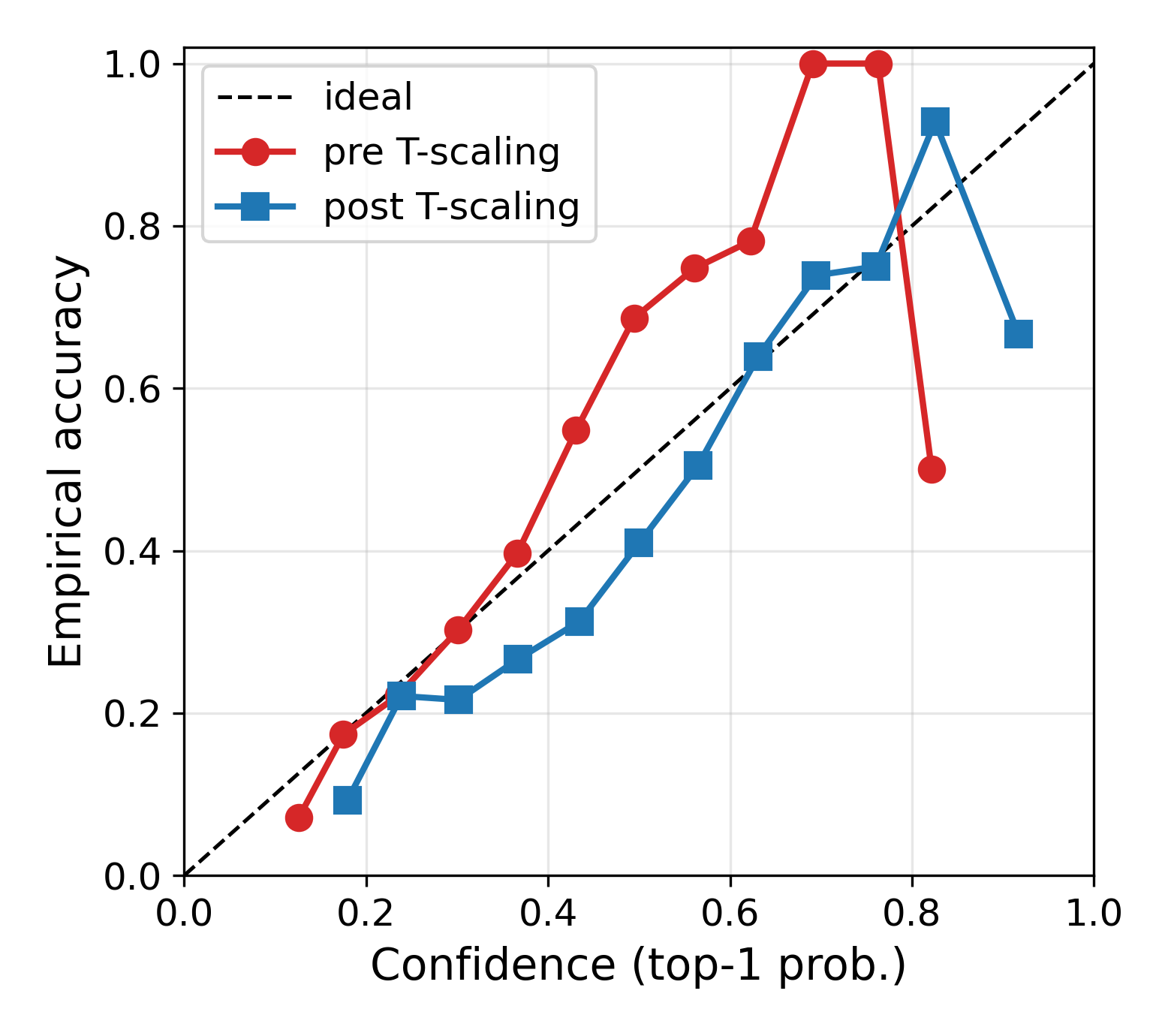}
\includegraphics[width=0.24\textwidth]{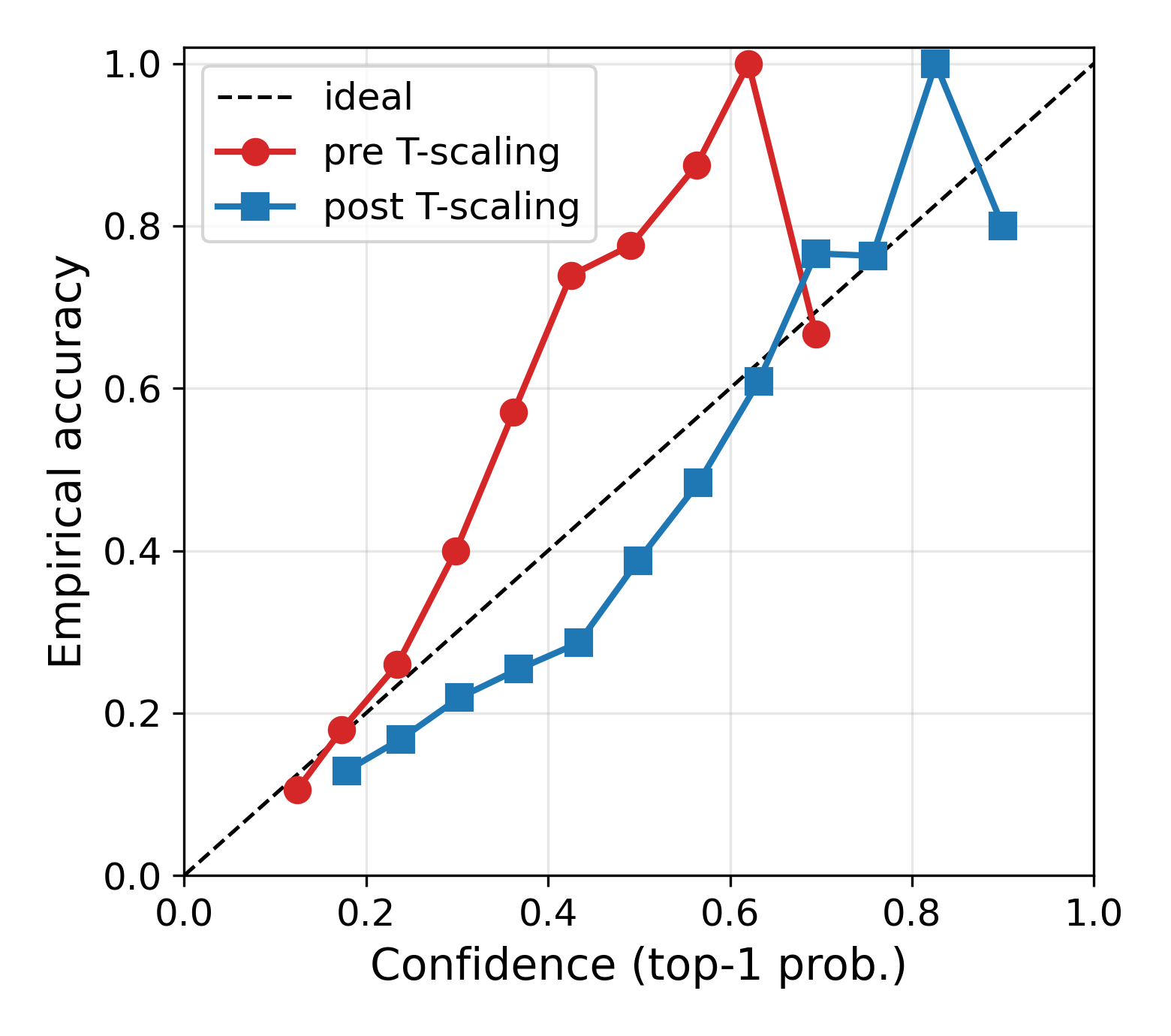}
\includegraphics[width=0.24\textwidth]{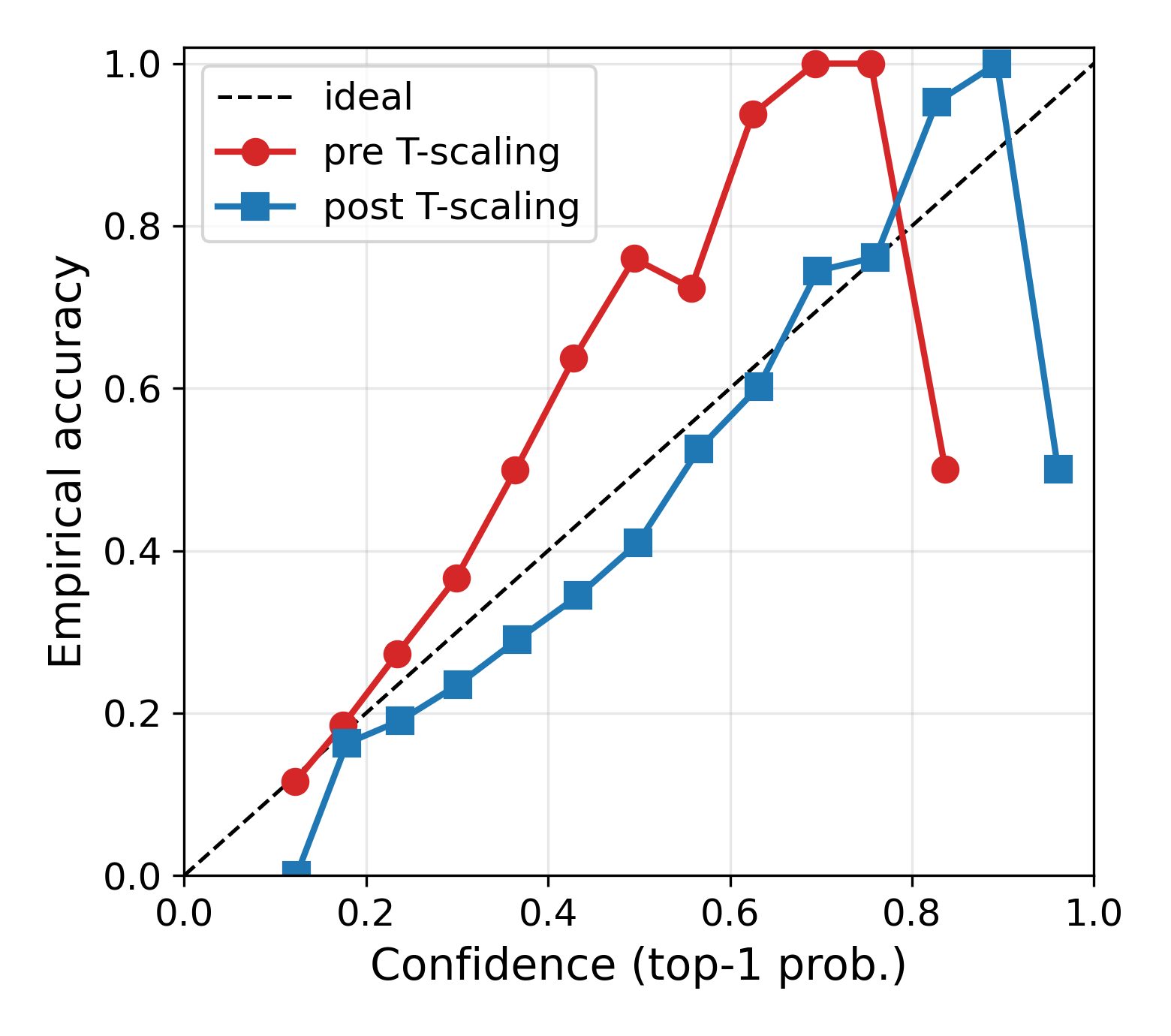}
\includegraphics[width=0.24\textwidth]{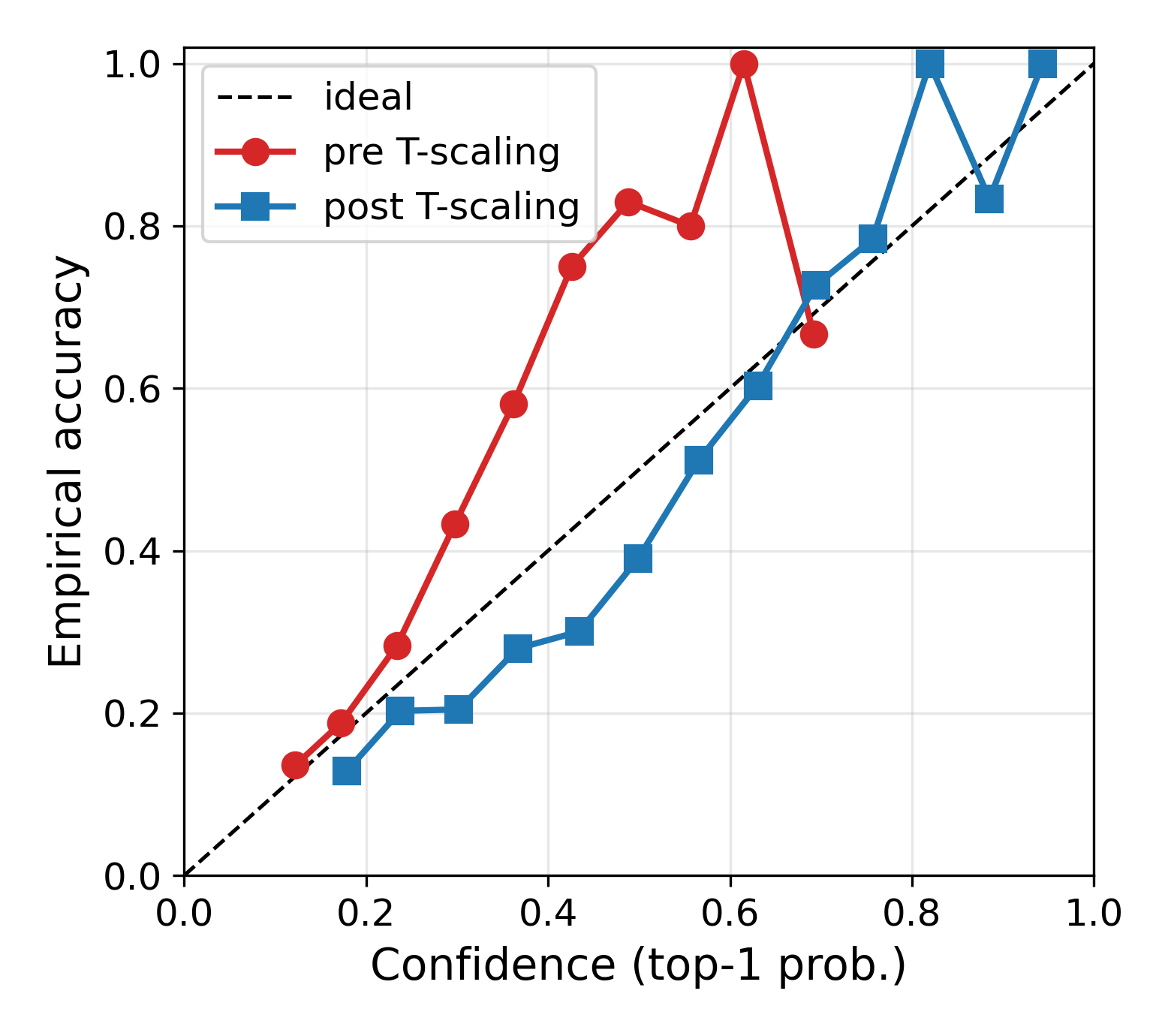}
\\
\includegraphics[width=0.24\textwidth]{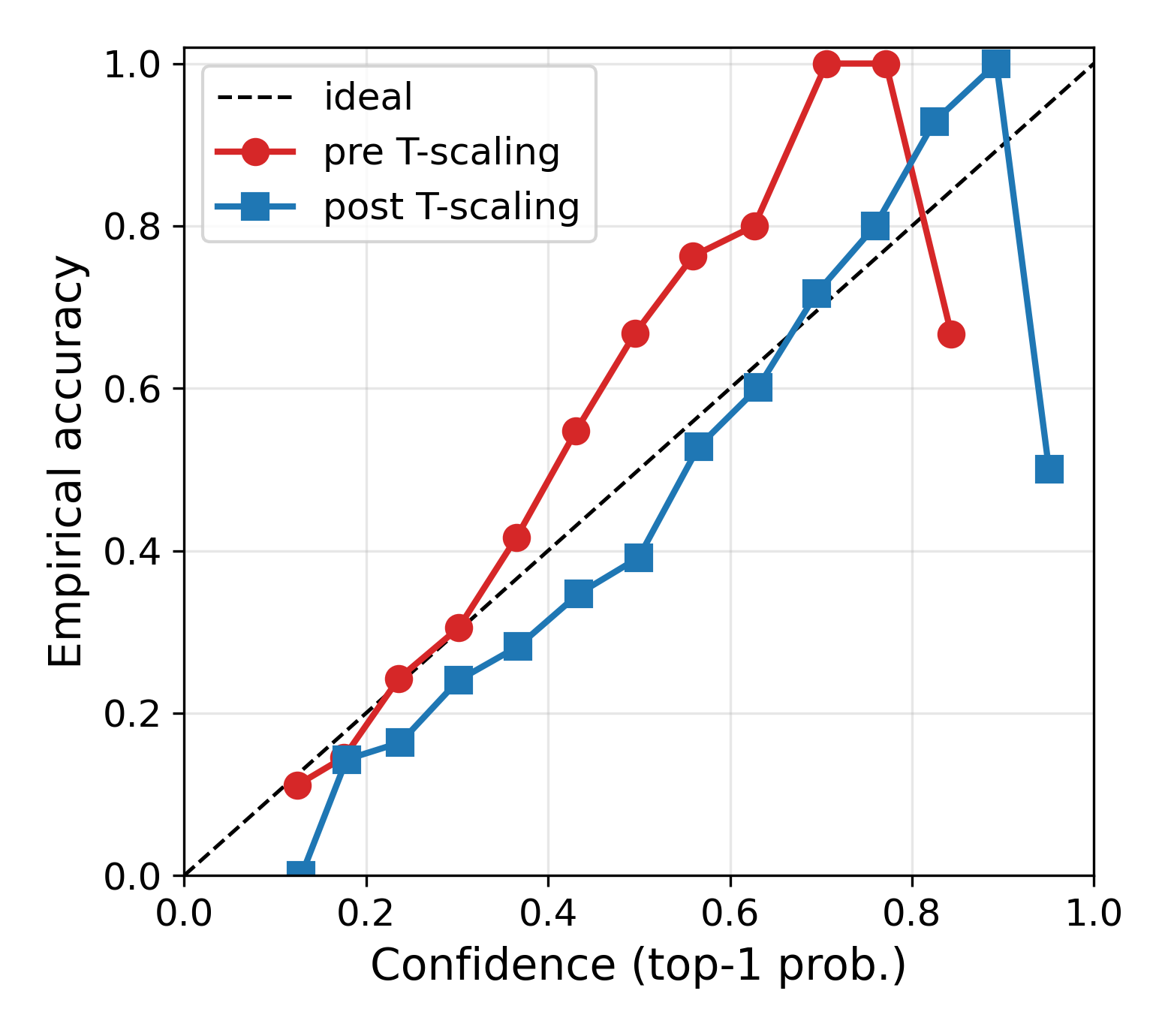}
\includegraphics[width=0.24\textwidth]{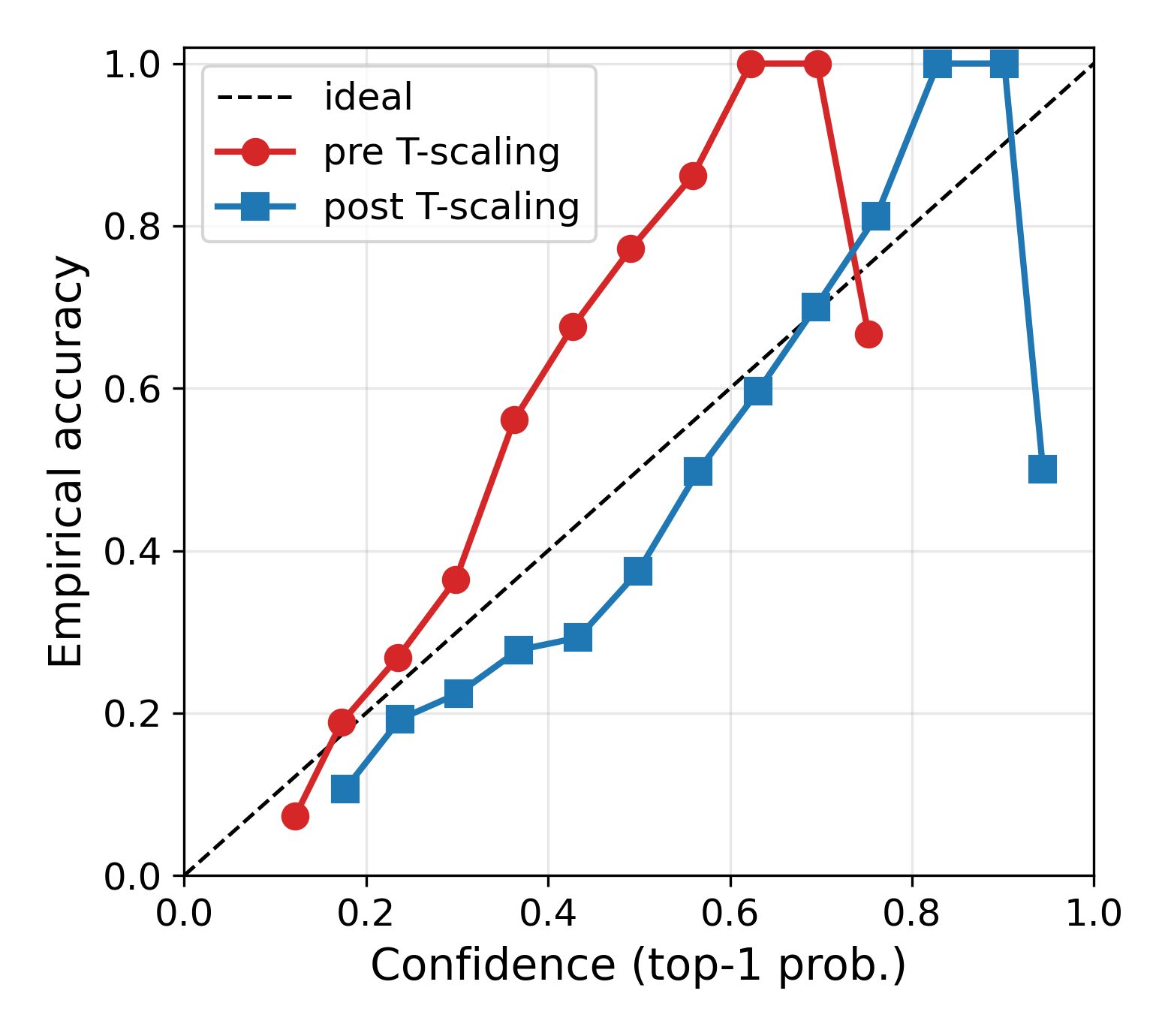}
\caption{Reliability diagrams for B0 / B1 / B2 (from left: B0\_hard, B0\_soft, B1\_hard, B1\_soft, B2\_hard, B2\_soft).}
\label{fig:reliability_baselines}
\end{figure}

The reliability shapes of the baselines differ qualitatively from those of the proposed method: the baselines concentrate in the high-confidence region ($> 0.7$), where empirical accuracy and confidence are distributed closely together, whereas the proposed method exhibits a posterior-tempered shape with a distribution that spreads across the low- and mid-confidence regions.

\section{Failure case analysis}\label{app:failure_full}

For examples that the proposed method gets wrong and that B2 (Deep Ensemble) gets right (the ``proposed wrong, B2 right'' quadrant), we analyse the top $30$ examples ranked by the proposed method's total entropy. The mean total entropy for the wrong examples is $2.87$ times that of the correct examples, indicating that the proposed method returns high entropy when it errs---a self-aware failure pattern. This provides quantitative support for the proposed method's uncertainty serving as a misclassification flag in reject-and-defer operation, and is consistent with the AURC / AUROC advantage on C5 in \S\ref{sec:c5_main} of the main body.

\section{Five-annotator high-disagreement subset}\label{app:5ann_subset}

This section collects the honest analysis on the high-disagreement subset referenced in \S\ref{sec:setup_data} of the main body ($1{,}544$ of $5{,}426$ validation examples; $28.5\%$). It is a group of examples to which a fifth annotator was added in the original GoEmotions data to resolve disagreement among the initial three raters.

On this subset, the JSD of the proposed method is higher than the baseline against B2 Deep Ensemble by $\Delta\,\jsd = +0.0205$ (paired $t_{(2)} = 25.82$, $p = 0.0015$; hard label, $3$ seeds). The diagnostic quantity $\kl(q \| p)$ on the same subset is $2.5$--$3.0$ for the proposed method and $1.34$--$1.41$ for the baselines, showing that the MCMC posterior mean assigns nearly zero probability to minority-vote categories. A hybrid head that combines posterior averaging with minority-vote smoothing is among the future-work directions mentioned in \S\ref{sec:disc_design_implications} and \S\ref{sec:limitations} of the main body.

\section{Computational cost and wall-clock breakdown}\label{app:wallclock}

The per-method comparison of training wall-clock is shown below (mean $\pm$ std over $6$ runs $= 2$ labels $\times$ $3$ seeds; NVIDIA H100 NVL).

{\small
\setlength{\tabcolsep}{4pt}
\begin{center}
\begin{tabular}{@{}l c c c@{}}
\toprule
Method & wall-clock (sec) & wall-clock (min) & ratio vs B0 \\
\midrule
B0 & $273.50 \pm 2.51$ & $4.56 \pm 0.04$ & $1.00\times$ \\
B1 (MC Dropout) & $709.00 \pm 3.46$ & $11.82 \pm 0.06$ & $2.59\times$ \\
B2 (Deep Ensemble) & $348.50 \pm 17.14$ & $5.81 \pm 0.29$ & $1.27\times$ \\
proposed & $243.33 \pm 2.58$& $4.06 \pm 0.04$& $0.89\times$\\
\bottomrule
\end{tabular}
\end{center}
}

The proposed method shows a $-11\%$ training-cost reduction relative to B0 and is more computationally efficient than B1 (MC Dropout, $20$ MCD passes) and B2 (Deep Ensemble, $5$ members). Inference time increases to $1.13\times$ B0 due to drawing $M = 30$ posterior samples, but because the frozen backbone restricts the $M$-fold computation to the head only, the actual latency stays at approximately $1.05\times$ the backbone forward pass.

The GPU-hour breakdown is approximately $14$ GPU-hours for the $24$ main-grid training runs, $4$ GPU-hours for the $9$ active-learning runs, $14$ GPU-hours for the $46$ robustness-ablation runs, and $22$ CPU-minutes for the $24$ post-hoc temperature-scaling runs (no GPU; single-CPU maximum-likelihood NLL), for a total of about $32$ GPU-hours ($+ 22$ CPU-minutes).

\end{document}